\documentclass{lovartreport}

\title{Folding Language Models into Themselves}
\author{Ziyang Zhang\textsuperscript{1,*,\ddag},\quad Yubin Jing\textsuperscript{2,*,\ddag},\quad Yuyao Li\textsuperscript{3,*},\quad Haofan Wang\textsuperscript{4},\quad Yichen Gong\textsuperscript{4,\dag}}
\reportbrand{Lovart}{Research}
\reporttype{TECHNICAL REPORT}
\reportnumber{02}
\reportversion{}
\reportshorttitle{Folding Language Models into Themselves}
\reportauthornote{\textsuperscript{1}Peking University\quad \textsuperscript{2}Georgia Institute of Technology\quad \textsuperscript{3}Tsinghua University\quad \textsuperscript{4}Lovart AI}
\reportabstractstyle{flat}
\reportmascot[20mm]{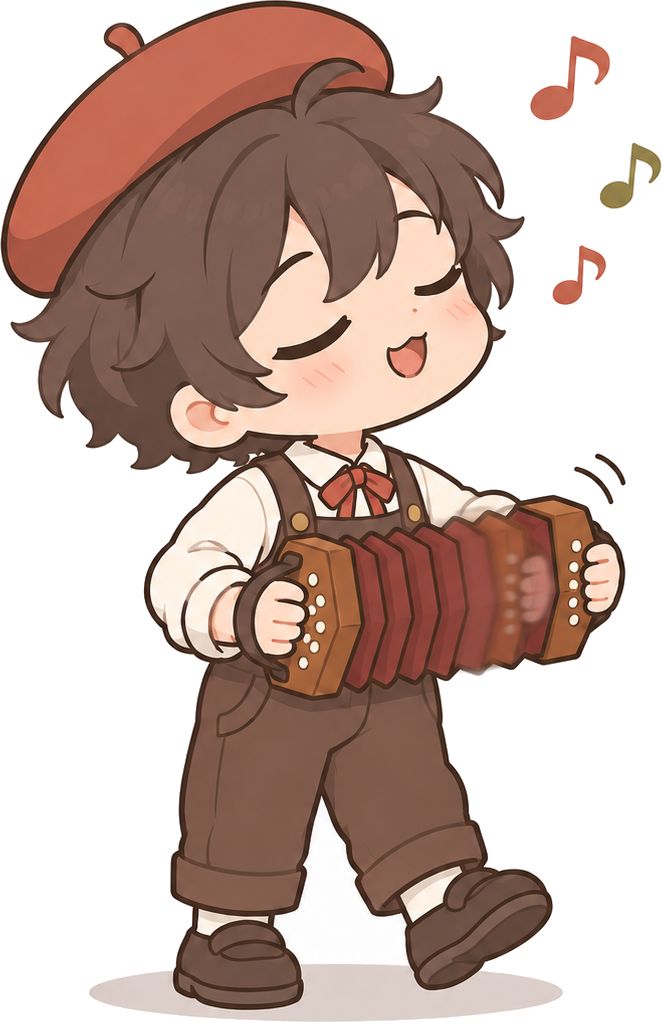}
\reportmascotoverlap{6mm}
\reportmascotxshift{-8mm}
\reportpartners{%
  \partnerwordmark{28mm}{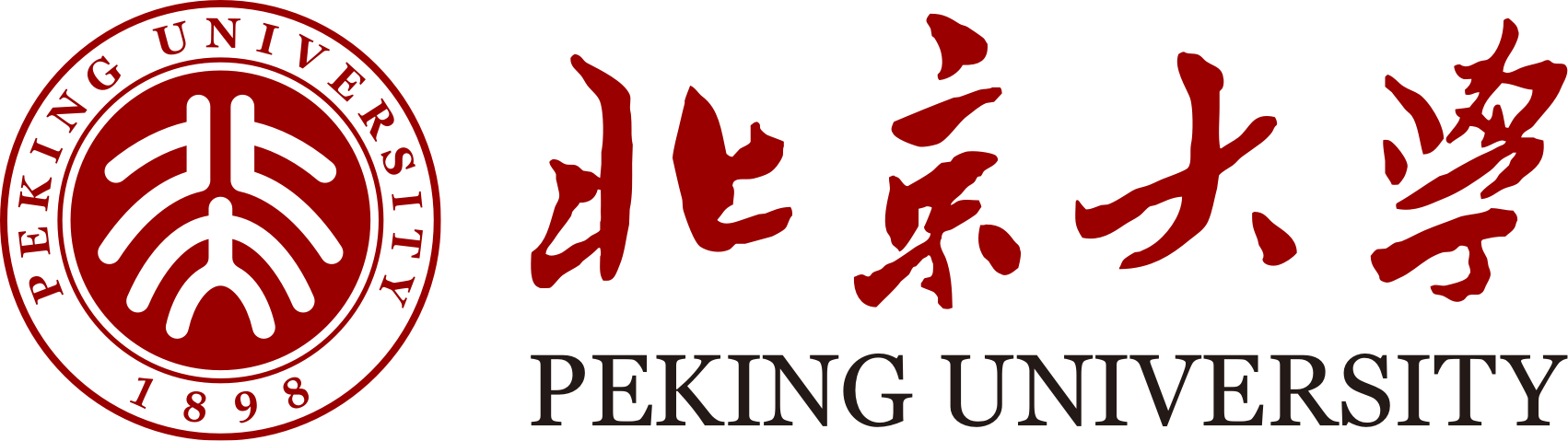}\hfill
  \partnerwordmark[trim=21bp 8bp 21bp 8bp,clip]{31mm}{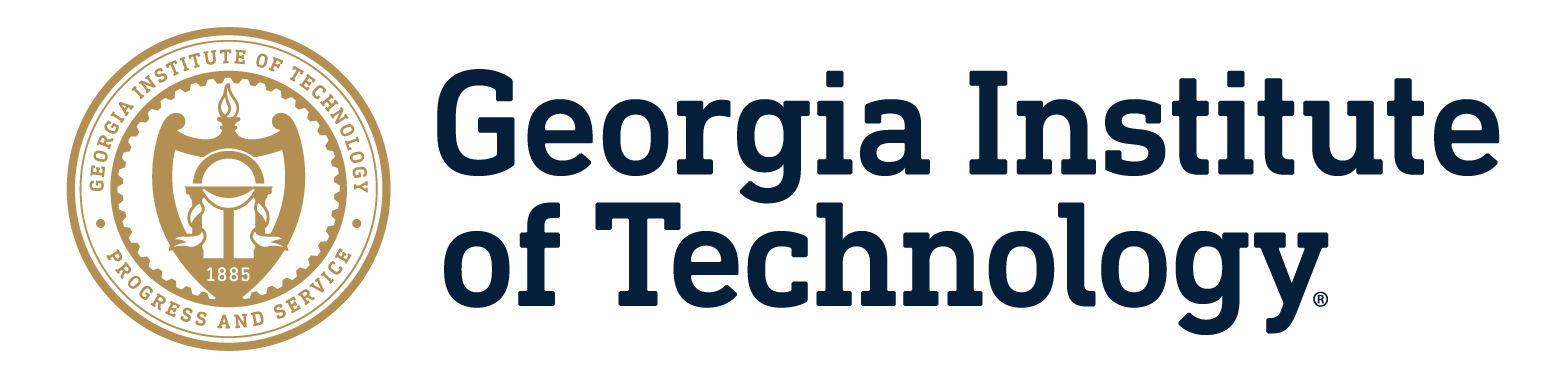}\hfill
  \partnerwordmark[trim=228bp 142bp 275bp 151bp,clip]{25mm}{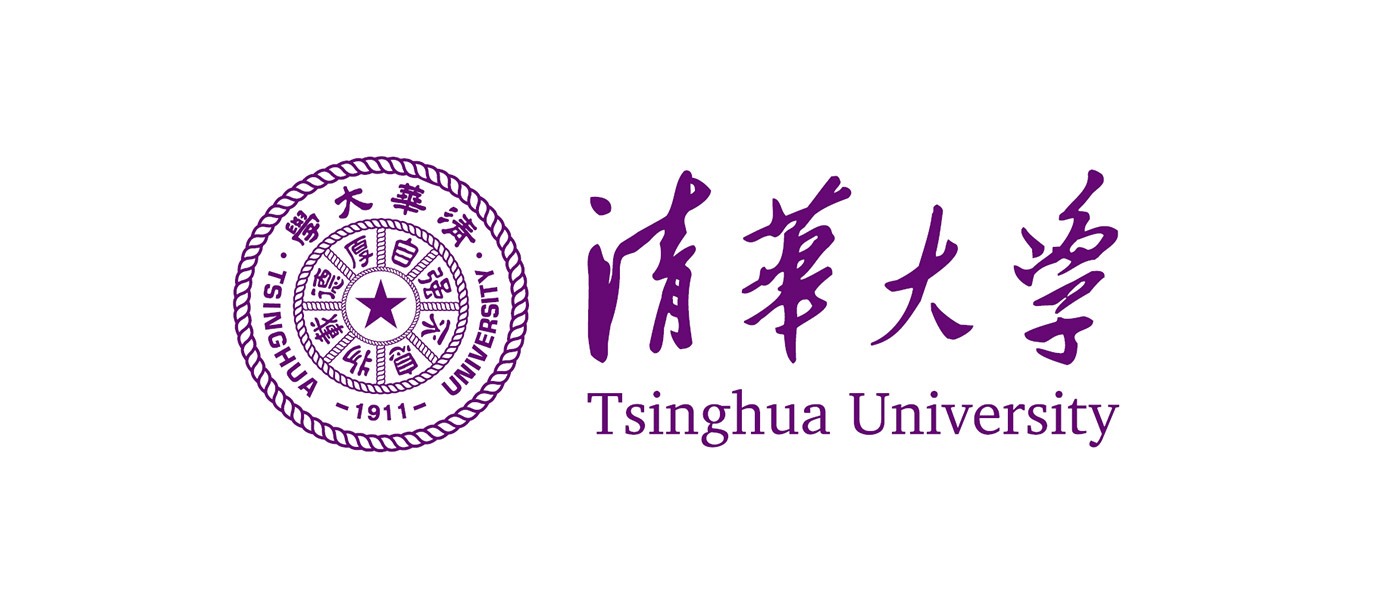}}
\date{November 2026}
\reportlinks{\href{https://www.lovart.ai}{lovart.ai}}
\hypersetup{pdftitle={Folding Language Models into Themselves},
  pdfauthor={Ziyang Zhang, Yubin Jing, Yuyao Li, Haofan Wang, Yichen Gong}}

\usepackage{placeins}
\makeatletter
\def\@makecol{\ifvoid\footins \setbox\@outputbox\box\@cclv
  \else \setbox\@outputbox\vbox{\boxmaxdepth\@maxdepth \unvbox\@cclv}\fi
  \let\@elt\relax \xdef\@freelist{\@freelist\@midlist}%
  \global\let\@midlist\@empty
  \@combinefloats
  \ifvoid\footins\else
    \setbox\@outputbox\vbox{\boxmaxdepth\@maxdepth \unvbox\@outputbox
      \vskip\skip\footins \color@begingroup\normalcolor\footnoterule
      \unvbox\footins \color@endgroup}\fi
  \ifvbox\@kludgeins \@makespecialcolbox
  \else \setbox\@outputbox\vbox to\@colht{\@texttop \dimen@\dp\@outputbox
    \unvbox\@outputbox \vskip-\dimen@ \@textbottom}\fi
  \global\maxdepth\@maxdepth}
\makeatother
\definecolor{ArchFG}{HTML}{6F8FAF}
\begin{document}
\maketitle
{\renewcommand{\thefootnote}{}\footnotetext{\textsuperscript{\ddag}Equal contribution.\quad \textsuperscript{*}Work done during an internship at Lovart AI.\quad \textsuperscript{\dag}Corresponding author.}}

\begin{abstract}
Knowledge distillation usually trains a compact student from scratch, which needs a large amount of data even though the teacher's weights already encode most of what the student has to learn.
In this work, we explore building the student directly from the teacher's weights.
We propose \textbf{Concertina}, which folds every group of $N$ adjacent teacher layers into one shared block that is executed $N$ times.
Each loop step carries a low-rank correction initialized from the layer it replaces, and a per-token gate mixes these corrections, to let every token recover the behavior of the original layers from a single shared block.
We further design a distillation data recipe that balances general, mathematical, academic and targeted multiple-choice data, to recover all capabilities of the teacher together rather than one at the expense of another.
On Qwen3-4B, a 2.27B Concertina model distilled on 22B tokens retains about 90\% of the teacher on commonsense and mathematical reasoning and 84\% on average across eight benchmarks.
It outperforms same-size dense and layer-sharing baselines, as well as Qwen3-1.7B, which is trained on over a thousand times more data.
These results suggest that a compact model need not be retrained at scale, since folding the teacher into itself and distilling it on the right data recovers most of its capabilities.
Code is available at \href{https://github.com/myboker/concertina}{github.com/myboker/concertina}.
\end{abstract}

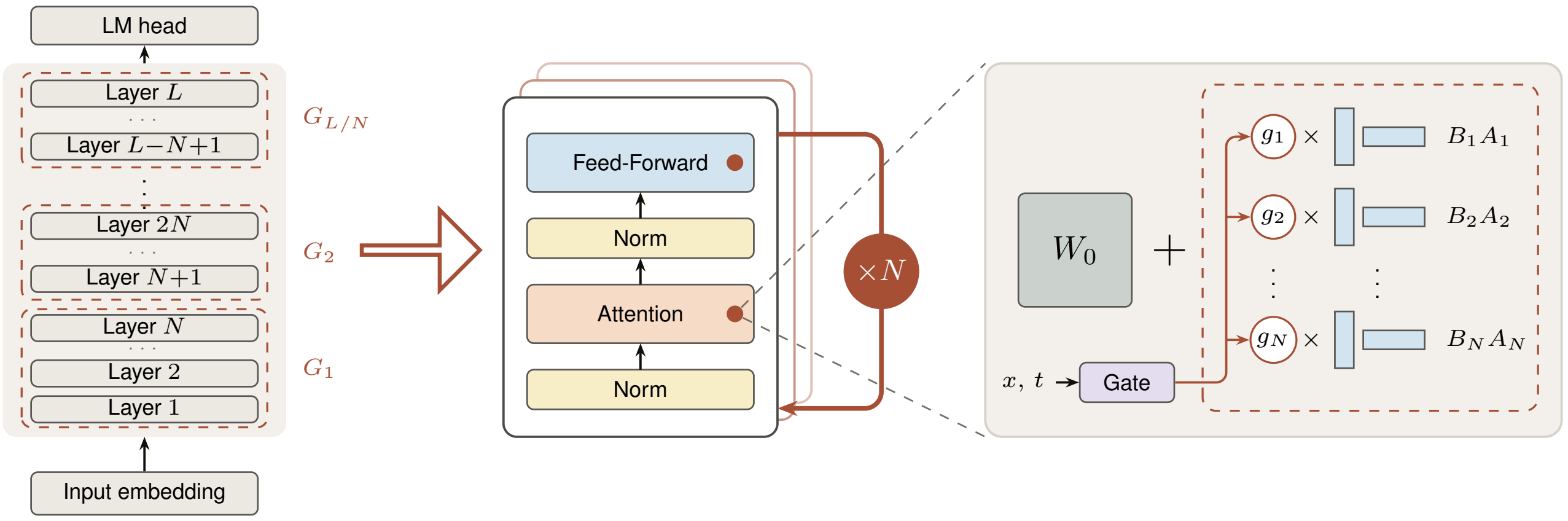
\begin{figure}[b!]
  \centering
\definecolor{ArchAttn}{HTML}{F6DCC6}
\definecolor{ArchFFN}{HTML}{D3E4F1}
\definecolor{ArchNorm}{HTML}{F7EDC6}
\definecolor{ArchGate}{HTML}{E4DDF0}
\definecolor{ArchIO}{HTML}{ECE8E2}
\usetikzlibrary{shapes.arrows}
\resizebox{\linewidth}{!}{%
\begin{tikzpicture}[font=\sffamily\scriptsize,
  box/.style={draw=LovartInk!70,line width=.5pt,rounded corners=2pt,align=center},
  layer/.style={box,fill=ArchIO,minimum width=2.4cm,minimum height=.28cm,inner sep=0pt},
  mod/.style={box,minimum width=2.4cm,inner sep=0pt},
  dot/.style={circle,fill=LovartAccent,inner sep=0pt,minimum size=5pt},
  arr/.style={-{Stealth[length=4pt]},line width=.7pt,draw=LovartInk},
  garr/.style={-{Stealth[length=4pt]},line width=.7pt,draw=LovartAccent}]

\fill[LovartLight,rounded corners=4pt] (-.1,.85) rectangle (2.9,4.8);
\node[box,fill=ArchIO,minimum width=2.4cm,minimum height=.36cm] (in) at (1.4,.25) {Input embedding};
\node[box,fill=ArchIO,minimum width=2.4cm,minimum height=.36cm] (out) at (1.4,5.2) {LM head};
\draw[arr] (in.north) -- (1.4,.85);
\draw[arr] (1.4,4.8) -- (out.south);
\draw[LovartAccent,dashed,line width=.6pt,rounded corners=3pt] (.1,.95) rectangle (2.7,2.2);
\node[layer] at (1.4,1.14) {Layer $1$};
\node[layer] at (1.4,1.52) {Layer $2$};
\node[text=LovartMuted,font=\tiny] at (1.4,1.78) {$\cdots$};
\node[layer] at (1.4,2.0) {Layer $N$};
\foreach \ya/\la/\lb in {2.3/{Layer $N{+}1$}/{Layer $2N$},3.7/{Layer $L{-}N{+}1$}/{Layer $L$}}{
  \draw[LovartAccent,dashed,line width=.6pt,rounded corners=3pt] (.1,\ya) rectangle (2.7,\ya+1.0);
  \node[layer] at (1.4,\ya+.22) {\la};
  \node[text=LovartMuted,font=\tiny] at (1.4,\ya+.5) {$\cdots$};
  \node[layer] at (1.4,\ya+.78) {\lb};}
\node[text=LovartAccent,anchor=west] at (2.95,1.575) {$G_1$};
\node[text=LovartAccent,anchor=west] at (2.95,2.8) {$G_2$};
\node[text=LovartAccent,anchor=west] at (2.95,4.2) {$G_{L/N}$};
\node at (1.4,3.52) {$\vdots$};

\node[single arrow,draw=LovartAccent,line width=1.2pt,fill=white,minimum height=1.25cm,minimum width=.85cm,single arrow head extend=.16cm,inner sep=0pt] at (4.3,2.825) {};

\foreach \d/\c in {.36/35,.18/60}{
  \draw[LovartAccent!\c,line width=.7pt,rounded corners=4pt,fill=white] (5.2+\d,.85+\d) rectangle (8.1+\d,4.44+\d);}
\draw[LovartInk!75,line width=.7pt,rounded corners=4pt,fill=white] (5.2,.85) rectangle (8.1,4.44);
\node[mod,fill=ArchNorm,minimum height=.42cm] at (6.65,1.35) {Norm};
\node[mod,fill=ArchAttn,minimum height=.62cm] (attn) at (6.65,2.15) {Attention};
\node[mod,fill=ArchNorm,minimum height=.42cm] at (6.65,2.95) {Norm};
\node[mod,fill=ArchFFN,minimum height=.62cm] (ffn) at (6.65,3.75) {Feed-Forward};
\draw[arr] (6.65,1.56) -- (6.65,1.84);
\draw[arr] (6.65,2.46) -- (6.65,2.74);
\draw[arr] (6.65,3.16) -- (6.65,3.44);
\node[dot] (fdot) at (7.65,2.15) {};
\node[dot] at (7.65,3.75) {};

\draw[LovartMuted,dashed,line width=.5pt] (fdot) -- (10.3,4.8);
\draw[LovartMuted,dashed,line width=.5pt] (fdot) -- (10.3,.85);

\draw[-{Stealth[length=6pt]},line width=1.4pt,LovartAccent,rounded corners=5pt]
  (8.1,4.05) -- (9.2,4.05) -- (9.2,1.15) -- (8.1,1.15);
\node[circle,fill=LovartAccent,text=white,inner sep=1pt,minimum size=.8cm,font=\sffamily\small\bfseries] at (9.2,2.6) {$\times N$};

\draw[LovartRule,line width=.7pt,rounded corners=5pt,fill=LovartLight] (10.3,.85) rectangle (16.4,4.8);
\draw[LovartInk!70,line width=.5pt,rounded corners=2pt,fill=LovartSage!55] (10.65,2.225) rectangle (11.85,3.425);
\node at (11.25,2.825) {\normalsize $W_0$};
\node at (12.25,2.825) {\Large $+$};
\draw[LovartAccent,dashed,line width=.6pt,rounded corners=4pt] (12.6,1.125) rectangle (16.15,4.575);
\foreach \yc/\k in {4.025/1,3.175/2,1.875/N}{
  \node[circle,draw=LovartAccent,fill=white,line width=.6pt,inner sep=1pt,minimum size=.46cm] (g\k) at (13.35,\yc) {$g_{\k}$};
  \node at (13.75,\yc) {$\times$};
  \draw[LovartInk!70,line width=.5pt,fill=ArchFFN] (14.0,\yc-.3) rectangle (14.2,\yc+.3);
  \draw[LovartInk!70,line width=.5pt,fill=ArchFFN] (14.3,\yc-.1) rectangle (14.95,\yc+.1);
  \node[anchor=west] at (15.05,\yc) {$B_{\k}A_{\k}$};}
\node at (13.35,2.575) {$\vdots$};
\node at (14.45,2.575) {$\vdots$};
\node[box,fill=ArchGate,minimum width=1.0cm,minimum height=.42cm] (gate) at (11.8,1.425) {Gate};
\node (xt) at (10.7,1.425) {$x,\,t$};
\draw[arr] (xt.east) -- (gate.west);
\draw[garr,rounded corners=3pt] (gate.east) -- (12.85,1.425) -- (12.85,4.025) -- (g1.west);
\draw[garr] (12.85,3.175) -- (g2.west);
\draw[garr] (12.85,1.875) -- (gN.west);
\end{tikzpicture}} {}
  \caption{Overview of Concertina.
    Left: the layers of the pretrained decoder are split into groups $G_1,\dots,G_{L/N}$ of $N$ adjacent layers.
    Middle: each group is replaced by one shared block that is executed $N$ times, and red dots mark the folded linear maps.
    Right: each folded linear map adds low-rank corrections $B_kA_k$ to the shared weight $W_0$, mixed by gate weights $g_k$ that depend on the token $x$ and the loop step $t$.
    Residual connections are omitted for clarity.}
  \label{fig:arch}
\end{figure}

\section{Introduction}
\label{sec:introduction}

Knowledge distillation is the standard way to obtain a compact language model from a larger teacher.
The student usually starts from random initialization and has to relearn from data everything the teacher already encodes.
Even the smallest members of recent open model families are therefore trained on trillions of tokens~\citep{gemma2024gemma2,yang2025qwen3}.
The cost of a compact model therefore lies in the data needed to rebuild it, not in its size.

A natural alternative is to build the student from the teacher's own weights.
Pruning whole layers is the most direct way, but it discards the computation of the removed layers and the pruned model loses much of its ability to generate.
\citet{bae2025rrt} and \citet{liu2024foldgpt} proposed to keep the depth instead, aiming to turn the teacher into a smaller model whose layers share weights.
Because such a student starts from the teacher, it recovers with tens of billions of tokens or fewer rather than the trillions needed from scratch.
However, existing designs either tie layers that behave differently or freeze the shared weights, and they adapt the shared block to each layer with a fixed update that is the same for every token.
The shared block then cannot play the roles of the layers it replaces, so the student starts far from the teacher and spends much of its data closing that gap.
What is missing is a way to share the teacher's weights that ties similar layers, lets each token adapt the shared computation, and trains every weight.

We propose Concertina, which folds every group of $N$ adjacent teacher layers into one shared block that is executed $N$ times.
We fold adjacent layers because they perform similar updates, much more so than the distant layers that are tied by layer sharing.
Each loop step carries a low-rank correction initialized from the layer it replaces, so the student starts close to the teacher.
A per-token gate mixes these corrections, so every token can decide how much the shared block acts as each of the original layers.
The folded student starts as an approximation of the teacher, so distillation only has to refine it.

During training, we found that the mixture of the distillation data, more than its amount, decides which capabilities the folded student recovers.
At the same budget of 22B tokens, a mixture that adds multiple-choice questions from many subjects and synthetic question-answer exercises raises the average recovery from 79\% to 84\% of the teacher, with the largest gains on MMLU-Pro, BBH and mathematics.
With this recipe, all eight benchmarks rise together as data grows.

On Qwen3-4B, a 2.27B Concertina model distilled on 22B tokens retains about 90\% of the teacher on commonsense and mathematical reasoning and 84\% on average across eight benchmarks, and outperforms Qwen3-1.7B, which is trained on over a thousand times more data.
Against a dense student of the same size distilled with the same data, it scores 8 to 11 points higher on the MMLU family and about 41 times higher on GSM8K.
Against the published recipes of prior weight-sharing designs of the same size, the full Concertina recipe on 197M tokens already exceeds both of them trained on ten times more data.
Under the same training recipe, adjacent folding and the per-token gate each give higher accuracy than the choices made by earlier designs.
A compact model need not be retrained at scale when it is folded from its teacher and distilled on the right data.

We summarize our contributions as follows.
\begin{itemize}
  \item Concertina, which folds adjacent teacher layers into looped shared blocks with per-token gated corrections.
  \item A distillation data recipe that recovers commonsense, knowledge and mathematical reasoning together, with an analysis of how each type of data contributes to each of them.
  \item A 2.27B model distilled from Qwen3-4B on 22B tokens, together with controlled comparisons against a same-size dense student and prior weight-sharing designs.
\end{itemize}
\section{Related Work}
\label{sec:related}

\paragraph{Knowledge distillation for language models.}
Knowledge distillation~\citep{hinton2015distilling} trains a student to match the output distribution of a teacher.
For language models it is applied to the next-token distribution on fixed text, to text generated by the teacher in sequence-level distillation~\citep{kim2016sequence}, and to intermediate representations, namely hidden states in DistilBERT~\citep{sanh2019distilbert} and both hidden states and attention maps in TinyBERT~\citep{jiao2020tinybert}.
On-policy distillation~\citep{agarwal2024policy,gu2024minillm} instead distills on the student's own generations, and it is now a common last stage of post-training~\citep{lu2025onpolicy,yang2025qwen3}.
Subliminal learning~\citep{cloud2026language} and active taskless distillation~\citep{zhang2026post} show that a student starting from the teacher's weights can learn from data unrelated to the target behavior.
These works study which signal the student learns from, whereas we study how to build the student from the teacher itself.

\paragraph{Parameter sharing across depth.}
Many works share weights across the layers of a transformer to reduce its parameters while keeping its depth.
Universal Transformers~\citep{dehghani2019universal}, ALBERT~\citep{lan2020albert}, MobileLLM~\citep{liu2024mobilellm} and Mixture-of-Recursions~\citep{bae2025mor} are trained from scratch with sharing in place, and Mixture of LoRAs~\citep{nouriborji2025mol} adds token-routed low-rank experts to such a model.
Relaxed Recursive Transformers~\citep{bae2025rrt} and FoldGPT~\citep{liu2024foldgpt} instead convert a pretrained model.
The first ties layer $i$ with layer $i+L/2$ and adds a static depth-wise LoRA to every loop, while the second removes redundant blocks and ties pairs of the remaining ones through learned scales on a frozen backbone.
DeltaLLM~\citep{mikaelyan2025deltallm}, FlexiGPT~\citep{smith2025flexigpt} and Ouroboros~\citep{jaber2026ouroboros} also share blocks with low-rank deltas, the last modulating them from the input.
Model folding~\citep{wang2025modelfolding} merges structurally similar neurons without data or fine-tuning.
Concertina folds adjacent decoder layers, aligns each low-rank correction with one original layer, mixes the corrections per token and trains every weight.

\paragraph{Depth pruning.}
Another line of work compresses a pretrained model by removing whole layers that contribute little to its output.
ShortGPT~\citep{men2024shortgpt} removes the layers whose input and output are most similar, layer pruning with healing~\citep{gromov2024unreasonable} removes a large fraction of the deeper layers followed by brief fine-tuning, and Minitron~\citep{muralidharan2024minitron} prunes depth and width and recovers by distillation.
These methods remove computation that the teacher relied on, whereas folding keeps every layer as one loop step of a shared block.
Section~\ref{sec:exp-dense} uses a distilled truncated model as the pruning control.

\paragraph{Low-rank adaptation.}
Low-rank updates change a large weight matrix with few trainable parameters.
LoRA~\citep{hu2022lora} adds a trainable low-rank update that starts at zero, and PiSSA~\citep{meng2024pissa} initializes it from the principal singular components of the pretrained weight.
Our step corrections are also low-rank and initialized by SVD, but they decompose the difference between each original layer and the shared weight rather than the weight itself, and the gate mixes them per token.

\section{Methodology}
\label{sec:method}

\paragraph{Overview.}
Concertina compresses a pretrained decoder by folding its layers onto themselves, as illustrated in Figure~\ref{fig:arch}.
It splits the $L$ layers into groups of $N$ adjacent layers and replaces each group with one shared block that is executed $N$ times.
This divides the weights of the layers by $N$, but asks one set of weights to play the role of $N$ different layers.
To let the shared block still act as each layer it replaces, every folded linear map adds low-rank step corrections to its shared weight.
The corrections are initialized from the original layers so that the folded model starts close to the teacher, and a per-token gate mixes them so that each token can choose how the block behaves at every step.
The folded model is then trained by distillation from the original model.

\subsection{Folding Adjacent Layers}
\label{sec:method-fold}
To reduce the parameters of the teacher without reducing its depth, we fold each group of adjacent layers into one block that is executed several times.
Folding with group size $N$ partitions the $L$ layers into $L/N$ groups of $N$ consecutive layers.
Each group is replaced by one block with the architecture of a single decoder layer, and the block runs $N$ times in a row.
We call each run a loop step.
Step $t = 1,\dots,N$ takes the place of the $t$-th layer of the group, and its output is the input of step $t+1$.
The folded model keeps the depth of the teacher but stores the linear weights of one layer per group.

For every linear map, the block keeps one shared weight $W_0$ that all $N$ steps use.
Let $W^{(1)},\dots,W^{(N)}$ be the weights of this map in the $N$ layers of the group, for each of the seven linear maps of a decoder layer, namely the query, key, value and output projections of attention and the gate, up and down projections of the feed-forward network.
We initialize the shared weight with the first layer,
\begin{equation}
  W_0 = W^{(1)},
\end{equation}
so the first step starts as an exact copy of that layer.
We anchor the shared weight to one layer instead of taking the mean of the group.
The first step then reproduces its layer exactly, and the whole difference within the group is left to the corrections of the remaining steps, so each correction has a single, well-defined target to learn.
We group adjacent rather than distant layers because consecutive layers of a pretrained model perform similar updates to the residual stream, so one shared weight can serve all of them with a small correction, as Section~\ref{sec:exp-design} confirms.

The RMSNorm layers are not shared, because they set the scale at which each layer reads its input.
Every RMSNorm of the block, including the query and key norms of Qwen3, keeps a separate scale vector for each step, initialized from the corresponding layer.
The embedding, the final norm and the language model head are not folded.
For Qwen3-4B with $L = 36$ and $N = 2$, all 36 layers are folded into 18 blocks, which gives 2.27B parameters or 57\% of the teacher, and $N = 3$ gives 12 blocks and 1.70B parameters.

\subsection{SVD-Initialized Step-LoRA}
\label{sec:method-lora}
A single shared weight cannot reproduce $N$ different layers, so we give every loop step its own low-rank correction, which aims to recover what distinguishes its layer from the shared weight.
For step $k$ and each folded linear map, the correction is a product $B_k A_k$ with $B_k \in \mathbb{R}^{d_{\mathrm{out}} \times r}$ and $A_k \in \mathbb{R}^{r \times d_{\mathrm{in}}}$, and step $k$ computes with the weight
\begin{equation}
  W_k = W_0 + B_k A_k .
\end{equation}

Instead of starting at zero as a LoRA update does, the correction starts from the actual difference between the original layer and the shared weight.
We take the truncated SVD $W^{(k)} - W_0 \approx U_r \Sigma_r V_r^{\top}$ and set
\begin{equation}
  B_k = U_r \Sigma_r^{1/2}, \qquad A_k = \Sigma_r^{1/2} V_r^{\top},
\end{equation}
so that $B_k A_k$ is the best rank-$r$ approximation of the difference.
The first step uses the shared weight directly and carries no correction, that is $B_1A_1 = 0$, and every later step starts at its own layer up to the top $r$ singular directions of the difference.
We use $r = 64$ for every folded linear map.
The folded model thus starts as an approximation of the teacher, and distillation only has to refine it rather than rebuild it.

\subsection{Input-Conditioned Step Gate}
\label{sec:method-gate}
A fixed correction per step treats every token in the same way, although different tokens may need the shared block to behave differently at the same step.
To let each token make this choice, we mix the corrections of all steps with a gate that depends on the hidden state of the token.
A folded linear map at step $t$ applies to an input $x \in \mathbb{R}^{d_{\mathrm{in}}}$ the weight
\begin{equation}
  W(x,t) = W_0 + \sum_{k=1}^{N} g_k(x,t)\, B_k A_k, \qquad g(x,t) = \mathrm{softmax}\big(W_g[t]^{\top} x + b[t]\big),
\end{equation}
where $W_g[t] \in \mathbb{R}^{d_{\mathrm{in}} \times N}$ and $b[t] \in \mathbb{R}^{N}$ are gate parameters specific to the step and to the linear map.
The output is computed as $W_0 x + \sum_k g_k(x,t)\, B_k (A_k x)$, so no full weight matrix is ever formed.

We initialize $W_g[t]$ to zero and $b[t]$ to a vector whose $t$-th entry is 8 and whose other entries are 0.
At the start of training $g(x,t)$ is therefore close to $\mathrm{onehot}(t)$ for every token, and the gated block matches the static corrections of Section~\ref{sec:method-lora}.
Training then decides how far each token departs from this assignment.
The static depth-wise LoRA of RRT is the special case in which $g$ stays a fixed one-hot vector.
With $N=2$ only the second step carries a correction, so the gate scales it for each token and places the block anywhere between the first layer and the approximated second layer at every step.
For the 18-block model the gates add 1.9M parameters, which is 0.08\% of the model.

\subsection{Distillation}
\label{sec:method-kd}
To recover what folding loses, we train the folded model to match the original model, which serves as a frozen teacher.
All parameters of the student are trained, including the shared weights, the step corrections, the norms, the gates, the embedding and the language model head.
We train the shared weights as well, so that each of them can move from a copy of one layer toward a common solution that serves every step of its group.
Every training step feeds the same batch to the teacher and the student, and at every position $i$ the loss combines the forward KL divergence from the teacher distribution $p_T$ to the student distribution $p_S$ with the cross entropy on the observed token,
\begin{equation}
  \mathcal{L}_i = \mathrm{KL}\big(p_T(\cdot \mid x_{<i}) \,\big\|\, p_S(\cdot \mid x_{<i})\big) - \lambda \log p_S(x_i \mid x_{<i}),
\end{equation}
with $\lambda = 0.25$ and temperature 1, averaged over all positions.
The KL term transfers the full output distribution of the teacher, and the cross entropy term keeps the student anchored to the data.
The dense student and the controlled RRT variants in Section~\ref{sec:experiments} use the same objective, so their differences from Concertina come from the structure alone.

\section{Experiments}
\label{sec:experiments}

This section first presents the main results of Concertina and how they improve with more distillation data.
It then compares Concertina with a same-size dense model and with prior weight-sharing designs, analyzes its design choices, studies the effect of the data mixture, and applies the method to a larger teacher.

\subsection{Experimental Setup}
\label{sec:setup}

\paragraph{Models.}
We use Qwen3-4B~\citep{yang2025qwen3} as the teacher in all experiments except Section~\ref{sec:exp-scale}, which uses Qwen3-8B.
The main student folds every two adjacent layers of the teacher into one shared block, that is $N=2$, and has 2.27B parameters.
We compare it with Qwen3-1.7B, the next smaller model of the same family, with a dense model truncated to the same size and distilled with the same recipe, and with RRT and FoldGPT reproduced on the same teacher.
All students start from the teacher weights, and Concertina and the dense control are trained with the objective of Section~\ref{sec:method-kd}.

\paragraph{Data.}
The main model is distilled on 22B tokens, of which 35\% are mathematical web text and solutions, 55\% are knowledge-oriented data and 10\% are general web text.
The knowledge-oriented part consists of multiple-choice questions from many subjects, diverse web text, and STEM and academic articles in a ratio of 25:18:12.
Part of the data is synthesized with Qwen3-30B-A3B, and Qwen3-235B-A22B cleans it by verifying each answer and checking its facts.
All data is decontaminated against every evaluation set by 13-gram matching.
We train with a peak learning rate of $3\times10^{-4}$, 500 warmup steps, cosine decay, sequences of 1024 tokens and about 295K tokens per step.
Section~\ref{sec:exp-mix} compares this mixture with two simpler ones at the same budget.

\paragraph{Benchmarks.}
We evaluate on eight benchmarks that range from saturated commonsense to hard knowledge and multi-step math.
The commonsense score, denoted base-7, is the mean of ARC-Challenge, ARC-Easy~\citep{clark2018think}, HellaSwag~\citep{zellers2019hellaswag}, PIQA~\citep{bisk2020piqa}, OpenBookQA~\citep{mihaylov2018can}, WinoGrande~\citep{sakaguchi2020winogrande} and LAMBADA~\citep{paperno2016lambada}, with per-task scores in Appendix~\ref{app:base7}.
These seven tasks, MMLU~\citep{hendrycks2021mmlu}, MMLU-Redux~\citep{gema2025mmluredux}, MMLU-Pro~\citep{wang2024mmlupro}, GPQA-diamond~\citep{rein2024gpqa} and BBH~\citep{suzgun2023bbh} are scored zero-shot by log-likelihood on their full test sets.
GSM8K~\citep{cobbe2021gsm8k} and MATH~\citep{hendrycks2021math} are scored by exact match on few-shot greedy generation, with MATH on a subset of 1{,}250 problems that takes every fourth problem of each subject.
We report each benchmark separately together with its percentage of the teacher score, because an average hides the gap on the hardest ones.

\subsection{Main Results}
\label{sec:exp-main}

\begin{table}[t]
  \centering
  \caption{Main results on eight benchmarks.
    The last row gives the score of Concertina as a percentage of the Qwen3-4B teacher.
    Tokens are pretraining tokens for the Qwen3 models and distillation tokens for Concertina, and bold marks the better of the two compact models.}
  \label{tab:main}
  \small
  \setlength{\tabcolsep}{3.6pt}
  \renewcommand{\arraystretch}{1.15}
  \begin{tabularx}{\linewidth}{@{}Xrrrrrrrrrr@{}}
    \toprule
    \textbf{Model} & \textbf{Params} & \textbf{Tokens} & \textbf{base-7} & \textbf{MMLU} &
      \textbf{Redux} & \textbf{Pro} & \textbf{GPQA} & \textbf{BBH} &
      \textbf{GSM8K} & \textbf{MATH}\\
    \midrule
    Qwen3-4B (teacher) & 4.02B & 36T & 0.630 & 0.671 & 0.720 & 0.404 & 0.379 & 0.505 & 0.879 & 0.566\\
    Qwen3-1.7B & 1.72B & 36T & 0.562 & \textbf{0.557} & \textbf{0.591} & 0.297 & 0.293 & 0.363 & 0.710 & 0.460\\
    \rowcolor{LovartLight}
    \textbf{Concertina} & 2.27B & 22B & \textbf{0.580} & 0.556 & 0.587 & \textbf{0.301} & \textbf{0.308} & \textbf{0.400} & \textbf{0.788} & \textbf{0.498}\\
    \midrule
    \% of teacher & 57\% & & 92\% & 83\% & 82\% & 75\% & 81\% & 79\% & 90\% & 88\%\\
    \bottomrule
  \end{tabularx}
\end{table}

Table~\ref{tab:main} compares Concertina with its teacher and with Qwen3-1.7B, the closest model of the same family in size.
With 57\% of the teacher's parameters, Concertina keeps 92\% of its base-7 score, 90\% of its GSM8K score and 88\% of its MATH score, and 84\% on average over the eight benchmarks.
It scores higher than Qwen3-1.7B on six of the eight benchmarks and stays within 0.4 points on MMLU and MMLU-Redux.
Qwen3-1.7B is pretrained on about 36T tokens, while Concertina is distilled on 22B, more than a thousand times fewer.
Starting from the teacher's weights therefore yields a compact model of this quality from a small fraction of the usual training data.

\subsection{Data Scaling}
\label{sec:exp-data}

\begin{figure*}[t]
  \centering
  \resizebox{0.85\linewidth}{!}{%
\begin{tikzpicture}[font=\sffamily\scriptsize,
  grid/.style={draw=LovartRule,line width=.3pt},
  lab/.style={text=LovartMuted}]
\definecolor{S0}{HTML}{2a78d6}
\definecolor{S1}{HTML}{eb6834}
\definecolor{S2}{HTML}{1baf7a}
\definecolor{S3}{HTML}{eda100}
\definecolor{S4}{HTML}{e87ba4}
\definecolor{S5}{HTML}{008300}
\definecolor{S6}{HTML}{6250d6}
\definecolor{S7}{HTML}{e34948}
\draw[grid] (0,0.000)--(11.000,0.000);
\node[anchor=east,lab] at (-.05,0.000){0.2};
\draw[grid] (0,0.829)--(11.000,0.829);
\node[anchor=east,lab] at (-.05,0.829){0.3};
\draw[grid] (0,1.657)--(11.000,1.657);
\node[anchor=east,lab] at (-.05,1.657){0.4};
\draw[grid] (0,2.486)--(11.000,2.486);
\node[anchor=east,lab] at (-.05,2.486){0.5};
\draw[grid] (0,3.314)--(11.000,3.314);
\node[anchor=east,lab] at (-.05,3.314){0.6};
\draw[grid] (0,4.143)--(11.000,4.143);
\node[anchor=east,lab] at (-.05,4.143){0.7};
\draw[grid] (0,4.971)--(11.000,4.971);
\node[anchor=east,lab] at (-.05,4.971){0.8};
\draw[grid] (0,5.800)--(11.000,5.800);
\node[anchor=east,lab] at (-.05,5.800){0.9};
\draw[LovartMuted,line width=.4pt] (0,0)--(11.000,0);
\draw[LovartMuted,line width=.4pt] (0.000,0)--(0.000,-.06);
\node[anchor=north,lab] at (0.000,-.06){0};
\draw[LovartMuted,line width=.4pt] (2.500,0)--(2.500,-.06);
\node[anchor=north,lab] at (2.500,-.06){5};
\draw[LovartMuted,line width=.4pt] (5.000,0)--(5.000,-.06);
\node[anchor=north,lab] at (5.000,-.06){10};
\draw[LovartMuted,line width=.4pt] (7.500,0)--(7.500,-.06);
\node[anchor=north,lab] at (7.500,-.06){15};
\draw[LovartMuted,line width=.4pt] (10.000,0)--(10.000,-.06);
\node[anchor=north,lab] at (10.000,-.06){20};
\draw[LovartMuted,line width=.4pt] (11.000,0)--(11.000,-.06);
\node[anchor=north,lab] at (11.000,-.06){22};
\node[lab] at (5.500,-.55){Distillation tokens (B)};
\node[rotate=90,lab] at (-.75,2.900){Score};
\draw[S0,line width=1pt,line join=round] (1.100,2.627)--(2.200,2.676)--(3.300,2.859)--(4.400,2.958)--(5.500,3.016)--(6.600,3.115)--(7.700,3.115)--(8.800,3.140)--(9.900,3.132)--(11.000,3.149);
\fill[S0] (1.100,2.627) circle (1.3pt);
\fill[S0] (2.200,2.676) circle (1.3pt);
\fill[S0] (3.300,2.859) circle (1.3pt);
\fill[S0] (4.400,2.958) circle (1.3pt);
\fill[S0] (5.500,3.016) circle (1.3pt);
\fill[S0] (6.600,3.115) circle (1.3pt);
\fill[S0] (7.700,3.115) circle (1.3pt);
\fill[S0] (8.800,3.140) circle (1.3pt);
\fill[S0] (9.900,3.132) circle (1.3pt);
\fill[S0] (11.000,3.149) circle (1.3pt);
\draw[S0,line width=.6pt,dash pattern=on 3pt off 2pt] (0,3.563)--(11.000,3.563);
\draw[S1,line width=1pt,line join=round] (1.100,2.171)--(2.200,2.287)--(3.300,2.577)--(4.400,2.709)--(5.500,2.817)--(6.600,2.834)--(7.700,2.908)--(8.800,2.950)--(9.900,2.966)--(11.000,2.950);
\fill[S1] (1.100,2.171) circle (1.3pt);
\fill[S1] (2.200,2.287) circle (1.3pt);
\fill[S1] (3.300,2.577) circle (1.3pt);
\fill[S1] (4.400,2.709) circle (1.3pt);
\fill[S1] (5.500,2.817) circle (1.3pt);
\fill[S1] (6.600,2.834) circle (1.3pt);
\fill[S1] (7.700,2.908) circle (1.3pt);
\fill[S1] (8.800,2.950) circle (1.3pt);
\fill[S1] (9.900,2.966) circle (1.3pt);
\fill[S1] (11.000,2.950) circle (1.3pt);
\draw[S1,line width=.6pt,dash pattern=on 3pt off 2pt] (0,3.903)--(11.000,3.903);
\draw[S2,line width=1pt,line join=round] (1.100,2.395)--(2.200,2.486)--(3.300,2.834)--(4.400,2.983)--(5.500,3.115)--(6.600,3.115)--(7.700,3.157)--(8.800,3.240)--(9.900,3.215)--(11.000,3.207);
\fill[S2] (1.100,2.395) circle (1.3pt);
\fill[S2] (2.200,2.486) circle (1.3pt);
\fill[S2] (3.300,2.834) circle (1.3pt);
\fill[S2] (4.400,2.983) circle (1.3pt);
\fill[S2] (5.500,3.115) circle (1.3pt);
\fill[S2] (6.600,3.115) circle (1.3pt);
\fill[S2] (7.700,3.157) circle (1.3pt);
\fill[S2] (8.800,3.240) circle (1.3pt);
\fill[S2] (9.900,3.215) circle (1.3pt);
\fill[S2] (11.000,3.207) circle (1.3pt);
\draw[S2,line width=.6pt,dash pattern=on 3pt off 2pt] (0,4.309)--(11.000,4.309);
\draw[S3,line width=1pt,line join=round] (1.100,0.215)--(2.200,0.340)--(3.300,0.481)--(4.400,0.563)--(5.500,0.688)--(6.600,0.721)--(7.700,0.804)--(8.800,0.820)--(9.900,0.829)--(11.000,0.837);
\fill[S3] (1.100,0.215) circle (1.3pt);
\fill[S3] (2.200,0.340) circle (1.3pt);
\fill[S3] (3.300,0.481) circle (1.3pt);
\fill[S3] (4.400,0.563) circle (1.3pt);
\fill[S3] (5.500,0.688) circle (1.3pt);
\fill[S3] (6.600,0.721) circle (1.3pt);
\fill[S3] (7.700,0.804) circle (1.3pt);
\fill[S3] (8.800,0.820) circle (1.3pt);
\fill[S3] (9.900,0.829) circle (1.3pt);
\fill[S3] (11.000,0.837) circle (1.3pt);
\draw[S3,line width=.6pt,dash pattern=on 3pt off 2pt] (0,1.690)--(11.000,1.690);
\draw[S4,line width=.6pt,line join=round] (1.100,0.481)--(2.200,0.771)--(3.300,0.936)--(4.400,0.389)--(5.500,0.853)--(6.600,0.895)--(7.700,1.102)--(8.800,0.978)--(9.900,0.853)--(11.000,0.895);
\fill[S4] (1.100,0.481) circle (1.3pt);
\fill[S4] (2.200,0.771) circle (1.3pt);
\fill[S4] (3.300,0.936) circle (1.3pt);
\fill[S4] (4.400,0.389) circle (1.3pt);
\fill[S4] (5.500,0.853) circle (1.3pt);
\fill[S4] (6.600,0.895) circle (1.3pt);
\fill[S4] (7.700,1.102) circle (1.3pt);
\fill[S4] (8.800,0.978) circle (1.3pt);
\fill[S4] (9.900,0.853) circle (1.3pt);
\fill[S4] (11.000,0.895) circle (1.3pt);
\draw[S4,line width=.6pt,dash pattern=on 3pt off 2pt] (0,1.483)--(11.000,1.483);
\draw[S5,line width=1pt,line join=round] (1.100,0.829)--(2.200,1.085)--(3.300,1.243)--(4.400,1.309)--(5.500,1.375)--(6.600,1.574)--(7.700,1.641)--(8.800,1.607)--(9.900,1.616)--(11.000,1.657);
\fill[S5] (1.100,0.829) circle (1.3pt);
\fill[S5] (2.200,1.085) circle (1.3pt);
\fill[S5] (3.300,1.243) circle (1.3pt);
\fill[S5] (4.400,1.309) circle (1.3pt);
\fill[S5] (5.500,1.375) circle (1.3pt);
\fill[S5] (6.600,1.574) circle (1.3pt);
\fill[S5] (7.700,1.641) circle (1.3pt);
\fill[S5] (8.800,1.607) circle (1.3pt);
\fill[S5] (9.900,1.616) circle (1.3pt);
\fill[S5] (11.000,1.657) circle (1.3pt);
\draw[S5,line width=.6pt,dash pattern=on 3pt off 2pt] (0,2.527)--(11.000,2.527);
\draw[S6,line width=1pt,line join=round] (1.100,2.776)--(2.200,3.447)--(3.300,3.753)--(4.400,4.135)--(5.500,4.358)--(6.600,4.764)--(7.700,4.905)--(8.800,4.847)--(9.900,4.864)--(11.000,4.872);
\fill[S6] (1.100,2.776) circle (1.3pt);
\fill[S6] (2.200,3.447) circle (1.3pt);
\fill[S6] (3.300,3.753) circle (1.3pt);
\fill[S6] (4.400,4.135) circle (1.3pt);
\fill[S6] (5.500,4.358) circle (1.3pt);
\fill[S6] (6.600,4.764) circle (1.3pt);
\fill[S6] (7.700,4.905) circle (1.3pt);
\fill[S6] (8.800,4.847) circle (1.3pt);
\fill[S6] (9.900,4.864) circle (1.3pt);
\fill[S6] (11.000,4.872) circle (1.3pt);
\draw[S6,line width=.6pt,dash pattern=on 3pt off 2pt] (0,5.626)--(11.000,5.626);
\draw[S7,line width=1pt,line join=round] (1.100,0.787)--(2.200,1.309)--(3.300,1.740)--(4.400,1.839)--(5.500,2.105)--(6.600,2.245)--(7.700,2.428)--(8.800,2.337)--(9.900,2.303)--(11.000,2.469);
\fill[S7] (1.100,0.787) circle (1.3pt);
\fill[S7] (2.200,1.309) circle (1.3pt);
\fill[S7] (3.300,1.740) circle (1.3pt);
\fill[S7] (4.400,1.839) circle (1.3pt);
\fill[S7] (5.500,2.105) circle (1.3pt);
\fill[S7] (6.600,2.245) circle (1.3pt);
\fill[S7] (7.700,2.428) circle (1.3pt);
\fill[S7] (8.800,2.337) circle (1.3pt);
\fill[S7] (9.900,2.303) circle (1.3pt);
\fill[S7] (11.000,2.469) circle (1.3pt);
\draw[S7,line width=.6pt,dash pattern=on 3pt off 2pt] (0,3.033)--(11.000,3.033);
\fill[white,draw=LovartRule,line width=.3pt] (0.150,5.866) rectangle (6.800,4.826);
\draw[S6,line width=1pt] (0.300,5.676)--(0.700,5.676); \fill[S6] (0.500,5.676) circle (1.3pt);
\node[anchor=west,inner sep=1pt] at (0.750,5.676){GSM8K};
\draw[S0,line width=1pt] (0.300,5.346)--(0.700,5.346); \fill[S0] (0.500,5.346) circle (1.3pt);
\node[anchor=west,inner sep=1pt] at (0.750,5.346){Base-7};
\draw[S2,line width=1pt] (0.300,5.016)--(0.700,5.016); \fill[S2] (0.500,5.016) circle (1.3pt);
\node[anchor=west,inner sep=1pt] at (0.750,5.016){MMLU-Redux};
\draw[S1,line width=1pt] (2.500,5.676)--(2.900,5.676); \fill[S1] (2.700,5.676) circle (1.3pt);
\node[anchor=west,inner sep=1pt] at (2.950,5.676){MMLU};
\draw[S7,line width=1pt] (2.500,5.346)--(2.900,5.346); \fill[S7] (2.700,5.346) circle (1.3pt);
\node[anchor=west,inner sep=1pt] at (2.950,5.346){MATH};
\draw[S5,line width=1pt] (2.500,5.016)--(2.900,5.016); \fill[S5] (2.700,5.016) circle (1.3pt);
\node[anchor=west,inner sep=1pt] at (2.950,5.016){BBH};
\draw[S4,line width=.6pt] (4.700,5.676)--(5.100,5.676); \fill[S4] (4.900,5.676) circle (1.3pt);
\node[anchor=west,inner sep=1pt] at (5.150,5.676){GPQA};
\draw[S3,line width=1pt] (4.700,5.346)--(5.100,5.346); \fill[S3] (4.900,5.346) circle (1.3pt);
\node[anchor=west,inner sep=1pt] at (5.150,5.346){MMLU-Pro};
\draw[LovartMuted,line width=.6pt,dash pattern=on 3pt off 2pt] (4.700,5.016)--(5.100,5.016); \node[anchor=west,inner sep=1pt] at (5.150,5.016){Teacher};
\end{tikzpicture}
}
  \caption{Scores of the main model on the eight benchmarks during training.
    Each line follows one benchmark at every 2.2B distillation tokens.
    Dashed lines mark the scores of the Qwen3-4B teacher in the color of each benchmark.}
  \label{fig:scaling}
\end{figure*}
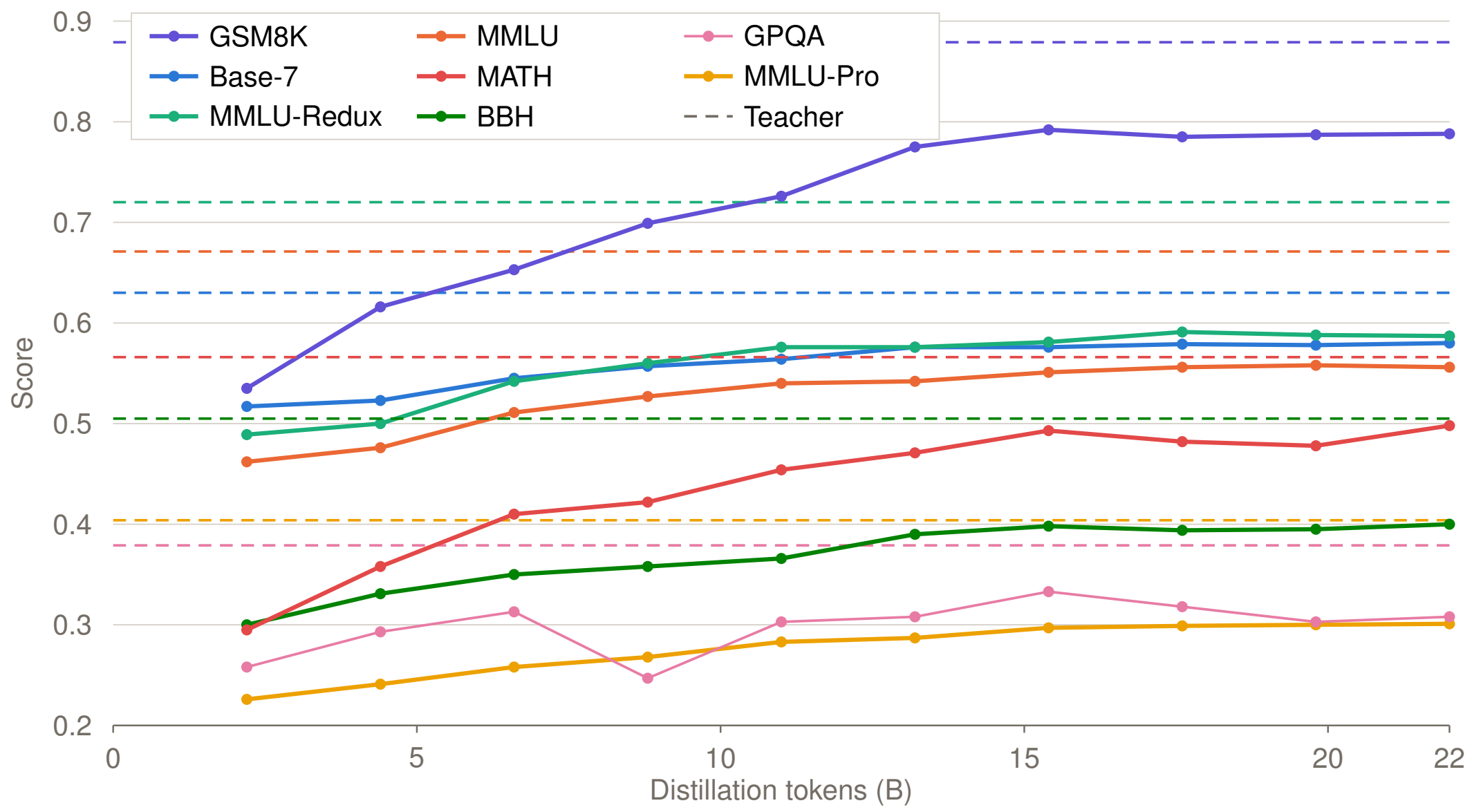

Figure~\ref{fig:scaling} shows how each of the eight benchmarks evolves over the training run of the main model.
All eight benchmarks rise together, and the average recovery grows from 64\% of the teacher at 2.2B tokens to 84\% at 15.4B tokens.
After that point the curves flatten, and most benchmarks change by less than one point over the last 7B tokens.
Apart from the noise of GPQA, which has only 198 questions, no benchmark falls while another rises, so the mixture improves every capability together rather than trading one for another.
Concertina passes Qwen3-1.7B on base-7, BBH and GSM8K by 11B tokens and on MATH by 13.2B tokens, and by 11B tokens its average recovery already exceeds that of the generic mixture after 22B tokens.

\subsection{Folding versus a Same-Recipe Dense Model}
\label{sec:exp-dense}

To test whether the folded structure adds value beyond the data, we compare it with a dense model of the same size trained under identical conditions.
The dense model keeps 19 of the 36 layers of Qwen3-4B, which gives 2.31B parameters without loops, step corrections or gates.
Both models start from Qwen3-4B and are distilled from it with the objective of Section~\ref{sec:method-kd} on the same 20B tokens of general, mathematical and academic text in a ratio of 25:35:40.

\begin{table}[htbp]
  \centering
  \caption{Folding versus a truncated dense model that keeps 19 of the 36 layers of Qwen3-4B.
    Under the same 20B-token distillation, the truncated model almost fails on GSM8K and MATH, while Concertina reaches 82\% and 72\% of the teacher.}
  \label{tab:dense}
  \small
  \setlength{\tabcolsep}{4.5pt}
  \renewcommand{\arraystretch}{1.15}
  \begin{tabularx}{\linewidth}{@{}Xrrrrrrrrr@{}}
    \toprule
    \textbf{Model} & \textbf{Params} & \textbf{base-7} & \textbf{MMLU} &
      \textbf{Redux} & \textbf{Pro} & \textbf{GPQA} & \textbf{BBH} &
      \textbf{GSM8K} & \textbf{MATH}\\
    \midrule
    Qwen3-4B (teacher) & 4.02B & 0.630 & 0.671 & 0.720 & 0.404 & 0.379 & 0.505 & 0.879 & 0.566\\
    Truncated dense & 2.31B & 0.570 & 0.434 & 0.453 & 0.157 & \textbf{0.298} & 0.289 & 0.017 & 0.038\\
    \rowcolor{LovartLight}
    \textbf{Concertina} & 2.27B & \textbf{0.590} & \textbf{0.533} & \textbf{0.566} & \textbf{0.238} & 0.283 & \textbf{0.332} & \textbf{0.718} & \textbf{0.408}\\
    \bottomrule
  \end{tabularx}
\end{table}

Table~\ref{tab:dense} compares the two students with the teacher on all eight benchmarks.
Concertina is higher on every multiple-choice benchmark except GPQA, by 8 to 11 points on the MMLU family and 4 points on BBH, and the 1.5-point gap on GPQA is within the noise of its 198 questions.
The difference is largest on generation, where Concertina scores about 41 times higher on GSM8K and 11 times higher on MATH.
The truncated model often computes the correct value but then repeats itself without stating an answer, so it keeps much of the teacher's knowledge but loses coherent generation.
At equal size, keeping the full depth of the teacher through loops is therefore worth more than keeping a subset of its layers.

\subsection{Prior Folding Methods}
\label{sec:exp-rrt}

Two prior methods also build a shared-weight student from the weights of a pretrained model.
\citet{bae2025rrt} proposed the Relaxed Recursive Transformer (RRT), which aims to convert a pretrained model into a looped one with few parameters.
It ties layer $i$ with layer $i + L/2$, loops the whole stack twice and attaches a static depth-wise LoRA to every loop.
\citet{liu2024foldgpt} proposed FoldGPT, which aims to compress a model by removing redundant blocks and sharing weights among the rest.
It first removes the least important blocks and then shares weights within pairs of adjacent remaining blocks, training only a small set of added parameters on a frozen backbone.
We compare Concertina with both methods trained with their published recipes.

\paragraph{Baseline protocol.}
All three methods compress Qwen3-4B into 18 unique blocks, which gives 2.27B parameters for Concertina, 2.34B for RRT and 2.21B for FoldGPT.
All three are trained on the same FineWeb-Edu text with sequences of 2048 tokens.
RRT follows the base uptraining recipe of its Appendix E, with average initialization, SVD-initialized rank-64 LoRA, tied norms and a cross-entropy loss.
FoldGPT follows its gate-based block removal and pairing, with a rank-8 LoRA on every parent block, trainable scales and norms, and a frozen backbone.
We give FoldGPT the same data, loss and learning rate as RRT instead of its original fine-tuning on 50K instructions.
All three methods are evaluated from 197M up to 2B tokens, with Concertina trained under the objective of Section~\ref{sec:method-kd}.
Further reproduction details are given in Appendix~\ref{app:repro}.
At these budgets the MMLU family stays close to chance for every method and cannot separate them.
We therefore report the base-7 average on at most 1{,}000 examples per task, on which the Qwen3-4B teacher scores 0.631, together with an eleven-task average that adds MMLU, MMLU-Redux, MMLU-Pro and SuperGPQA~\citep{du2025supergpqa}, and perplexity.

\begin{figure}[htbp]
  \centering
  \resizebox{\linewidth}{!}{%
\begin{tikzpicture}[font=\sffamily\scriptsize,
  grid/.style={draw=LovartRule,line width=.3pt},
  lab/.style={text=LovartMuted},
  rrt/.style={draw=gray!60,line width=1pt},
  fg/.style={draw=ArchFG,line width=1pt,dashed},
  ours/.style={draw=LovartAccent,line width=1.2pt}]
\definecolor{ArchFG}{HTML}{6F8FAF}

\begin{scope}
  \foreach \v/\y in {0.40/0.24,0.45/0.84,0.50/1.44,0.55/2.04,0.60/2.64}{
    \draw[grid] (0,\y)--(4.4,\y);
    \node[anchor=east,lab] at (-.05,\y){\v};}
  \draw[LovartMuted,line width=.4pt] (0,0)--(4.4,0);
  \foreach \x/\t in {0.4/197M,1.815/0.49B,2.923/1B,4.0/2B}{
    \node[anchor=north] at (\x,-.05){\t};}
  \node[lab] at (2.2,-.55){Training tokens};
  \node[rotate=90,lab] at (-.75,1.5){Base-7 accuracy};
  \draw[LovartInk!60,line width=.6pt,dashed] (0,3.0)--(4.4,3.0);
  \node[anchor=south east,text=LovartInk!70] at (4.4,3.0){Qwen3-4B 0.631};
  \draw[rrt] (0.4,0.42)--(1.815,0.924)--(2.923,1.14)--(4.0,1.2);
  \fill[gray!60] (0.4,0.42) circle (2.2pt) (1.815,0.924) circle (2.2pt) (2.923,1.14) circle (2.2pt) (4.0,1.2) circle (2.2pt);
  \draw[fg] (0.4,0.36)--(1.815,0.468)--(2.923,0.552)--(4.0,0.6);
  \fill[ArchFG] (0.4,0.36) circle (2.2pt) (1.815,0.468) circle (2.2pt) (2.923,0.552) circle (2.2pt) (4.0,0.6) circle (2.2pt);
  \draw[ours] (0.4,1.416)--(2.55,1.524)--(3.98,1.86);
  \fill[LovartAccent] (0.4,1.416) circle (2.6pt) (2.55,1.524) circle (2.6pt) (3.98,1.86) circle (2.6pt);
  \node[anchor=south,text=LovartAccent] at (0.4,1.47){0.498};
  \node[anchor=south,text=LovartAccent] at (3.98,1.91){0.535};
  \node[anchor=north,text=gray!80!black] at (4.0,1.17){0.480};
  \node[anchor=north,text=ArchFG] at (4.0,0.57){0.430};
  \node[font=\sffamily\small\bfseries] at (2.2,3.75){(a) Base-7};
\end{scope}

\begin{scope}[xshift=6.1cm]
  \foreach \v/\y in {0.32/0.3,0.34/0.9,0.36/1.5,0.38/2.1,0.40/2.7,0.42/3.3}{
    \draw[grid] (0,\y)--(4.4,\y);
    \node[anchor=east,lab] at (-.05,\y){\v};}
  \draw[LovartMuted,line width=.4pt] (0,0)--(4.4,0);
  \foreach \x/\t in {0.4/197M,1.815/0.49B,2.923/1B,4.0/2B}{
    \node[anchor=north] at (\x,-.05){\t};}
  \node[lab] at (2.2,-.55){Training tokens};
  \node[rotate=90,lab] at (-.75,1.5){Eleven-task average};
  \draw[rrt] (0.4,0.48)--(1.815,1.44)--(2.923,1.74)--(4.0,1.89);
  \fill[gray!60] (0.4,0.48) circle (2.2pt) (1.815,1.44) circle (2.2pt) (2.923,1.74) circle (2.2pt) (4.0,1.89) circle (2.2pt);
  \draw[fg] (0.4,0.51)--(1.815,0.51)--(2.923,0.6)--(4.0,0.75);
  \fill[ArchFG] (0.4,0.51) circle (2.2pt) (1.815,0.51) circle (2.2pt) (2.923,0.6) circle (2.2pt) (4.0,0.75) circle (2.2pt);
  \draw[ours] (0.4,2.25)--(2.55,2.43)--(3.98,2.97);
  \fill[LovartAccent] (0.4,2.25) circle (2.6pt) (2.55,2.43) circle (2.6pt) (3.98,2.97) circle (2.6pt);
  \node[anchor=south,text=LovartAccent] at (0.4,2.31){0.385};
  \node[anchor=south,text=LovartAccent] at (3.98,3.02){0.409};
  \node[anchor=north,text=gray!80!black] at (4.0,1.86){0.373};
  \node[anchor=north,text=ArchFG] at (4.0,0.72){0.335};
  \node[font=\sffamily\small\bfseries] at (2.2,3.75){(b) Eleven-task average};
\end{scope}

\begin{scope}[xshift=11.3cm,yshift=1.0cm]
  \draw[ours] (0,1.6)--(0.3,1.6); \fill[LovartAccent] (0.15,1.6) circle (2.6pt); \node[anchor=west] at (0.35,1.6){Concertina};
  \draw[rrt] (0,1.05)--(0.3,1.05); \fill[gray!60] (0.15,1.05) circle (2.2pt); \node[anchor=west] at (0.35,1.05){RRT};
  \draw[fg] (0,0.5)--(0.3,0.5); \fill[ArchFG] (0.15,0.5) circle (2.2pt); \node[anchor=west] at (0.35,0.5){FoldGPT};
\end{scope}
\end{tikzpicture}
}
  \caption{Accuracy of Concertina, RRT and FoldGPT against the number of training tokens.
    (a)~Base-7 average.
    (b)~Average over eleven tasks.
    Both baselines are trained with their published recipes on the same data, and the token axis is logarithmic.
    The middle point of Concertina is at 786M tokens, and the dashed line in (a) marks the Qwen3-4B teacher.}
  \label{fig:baselines}
\end{figure}
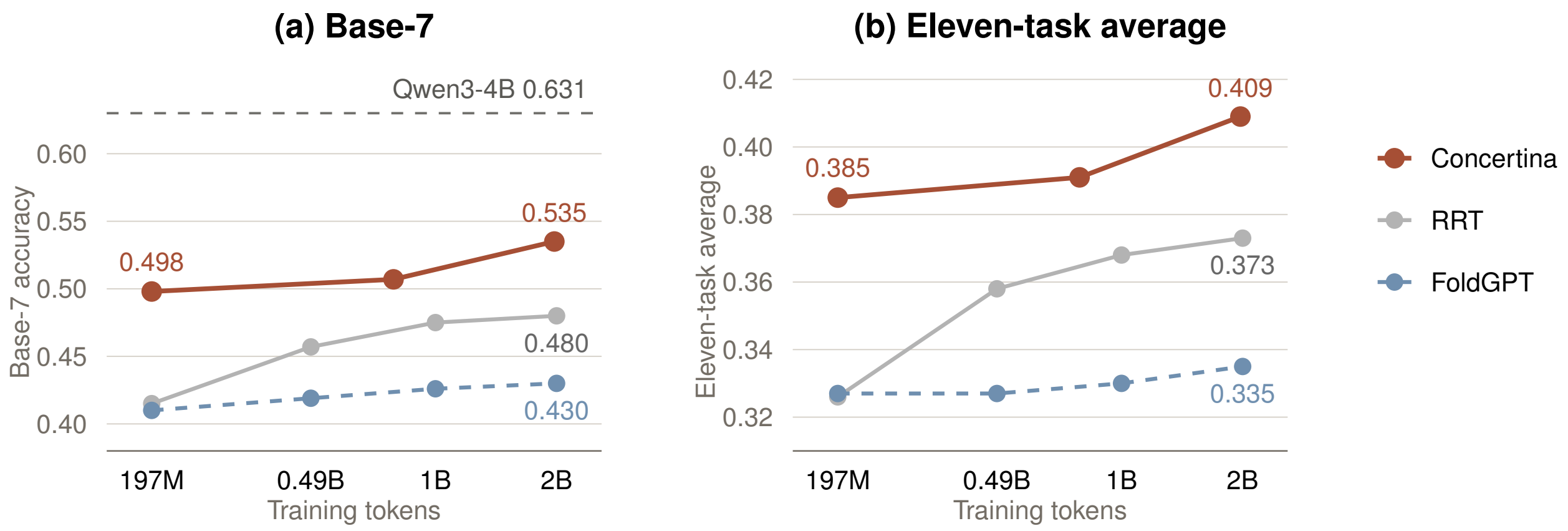

\paragraph{Data efficiency.}
Figure~\ref{fig:baselines} shows how the accuracy of the three methods grows with training data.
At an equal budget of 197M tokens, Concertina reaches 0.498 on base-7, ahead of RRT at 0.415 and FoldGPT at 0.410, and it also has the lowest perplexity of the three, as listed in Table~\ref{tab:baselines-full} of the appendix.
RRT improves steadily with more data but reaches only 0.480 at 2B tokens, which is still below Concertina trained on a tenth of that budget.
FoldGPT gains only 0.020 from 197M to 2B tokens and its perplexity stays near 20, because its backbone is frozen and only about 8.8M parameters are trained.
The eleven-task average gives the same ranking, with Concertina at 0.385 after 197M tokens against 0.373 for RRT and 0.335 for FoldGPT after 2B tokens.
Concertina also keeps improving with more data and reaches 0.535 on base-7 and 0.409 on the eleven-task average at 2B tokens, so it stays above both baselines at every budget.
As a complete recipe, Concertina therefore exceeds both published recipes with ten times less training data, and Section~\ref{sec:exp-design} isolates the part of this gap that comes from the structure.

\subsection{Analysis of Design Choices}
\label{sec:exp-design}

The published recipes differ from ours in more than structure, since RRT and FoldGPT are trained with cross-entropy while Concertina distills from the teacher.
To find which design choices account for the advantage over RRT, we hold the training recipe fixed and change one design axis at a time.
The first axis is the tying topology, that is, which layers share one block.
RRT ties layer $i$ with layer $i + L/2$, which we call cycle tying, whereas we tie each pair of adjacent layers.
The second axis is the per-loop modulation.
RRT adds a static depth-wise LoRA that is the same for every token, whereas our gate mixes the step corrections according to the input.
Every controlled run starts from Qwen3-4B, uses rank-64 step-LoRA with SVD initialization and step-specific RMSNorm, and is trained with the distillation objective of Section~\ref{sec:method-kd} under the same seed, batch size and learning rate.

\begin{figure}[htbp]
  \centering
\begin{tikzpicture}[font=\sffamily\scriptsize,
  rrt/.style={draw=gray!70,fill=gray!40},
  ours/.style={draw=LovartAccent,fill=LovartAccent},
  grid/.style={draw=LovartRule,line width=.3pt},
  lab/.style={text=LovartMuted}]

\begin{scope}
  \foreach \v/\y in {0.45/0.3,0.48/1.2,0.51/2.1,0.54/3.0}{
    \draw[grid] (0,\y)--(3.6,\y);
    \node[anchor=east,lab] at (-.05,\y){\v};}
  \draw[LovartMuted,line width=.4pt] (0,0)--(3.6,0);
  \node[rotate=90,lab] at (-.75,1.65){Base-7 accuracy};
  \node[anchor=north] at (0.8,-.05){197M};
  \node[anchor=north] at (2.8,-.05){2B};
  \node[lab] at (1.8,-.55){Distillation tokens};
  \draw[gray!60,line width=1pt] (0.8,0.84)--(2.8,2.43);
  \draw[LovartAccent,line width=1pt] (0.8,1.74)--(2.8,2.85);
  \fill[gray!60] (0.8,0.84) circle (2.2pt) (2.8,2.43) circle (2.2pt);
  \fill[LovartAccent] (0.8,1.74) circle (2.2pt) (2.8,2.85) circle (2.2pt);
  \node[anchor=north west,text=gray!80!black] at (0.85,0.82){0.468};
  \node[anchor=north west,text=gray!80!black] at (2.85,2.41){0.521};
  \node[anchor=south east,text=LovartAccent] at (0.78,1.76){0.498};
  \node[anchor=south east,text=LovartAccent] at (2.78,2.87){0.535};
  \node[font=\sffamily\small\bfseries] at (1.8,3.75){(a) Tying topology};
  \fill[gray!60] (0.1,3.32) circle (2pt); \node[anchor=west] at (0.18,3.32){Cycle (RRT)};
  \fill[LovartAccent] (2.0,3.32) circle (2pt); \node[anchor=west] at (2.08,3.32){Adjacent};
\end{scope}

\begin{scope}[xshift=5.15cm]
  \foreach \v/\y in {0.45/0,0.46/0.75,0.47/1.5,0.48/2.25,0.49/3.0}{
    \draw[grid] (0,\y)--(3.6,\y);
    \node[anchor=east,lab] at (-.05,\y){\v};}
  \draw[LovartMuted,line width=.4pt] (0,0)--(3.6,0);
  \node[anchor=north] at (0.8,-.05){197M};
  \node[anchor=north] at (2.8,-.05){490M};
  \node[lab] at (1.8,-.55){Distillation tokens};
  \draw[gray!60,line width=1pt] (0.8,0.375)--(2.8,1.575);
  \draw[LovartAccent,line width=1pt] (0.8,0.90)--(2.8,2.025);
  \fill[gray!60] (0.8,0.375) circle (2.2pt) (2.8,1.575) circle (2.2pt);
  \fill[LovartAccent] (0.8,0.90) circle (2.2pt) (2.8,2.025) circle (2.2pt);
  \node[anchor=north west,text=gray!80!black] at (0.85,0.35){0.455};
  \node[anchor=north west,text=gray!80!black] at (2.85,1.55){0.471};
  \node[anchor=south east,text=LovartAccent] at (0.78,0.92){0.462};
  \node[anchor=south east,text=LovartAccent] at (2.78,2.05){0.477};
  \node[font=\sffamily\small\bfseries] at (1.8,3.75){(b) Per-loop modulation};
  \fill[gray!60] (0.1,3.32) circle (2pt); \node[anchor=west] at (0.18,3.32){Static (RRT)};
  \fill[LovartAccent] (2.0,3.32) circle (2pt); \node[anchor=west] at (2.08,3.32){Gated};
\end{scope}

\begin{scope}[xshift=10.6cm]
  \foreach \v/\y in {0.0/0,0.1/0.6,0.2/1.2,0.3/1.8,0.4/2.4,0.5/3.0}{
    \draw[grid] (0,\y)--(3.2,\y);
    \node[anchor=east,lab] at (-.05,\y){\v};}
  \node[rotate=90,lab] at (-.7,1.5){Mean cosine};
  \fill[gray!60] (0.45,0) rectangle (1.35,1.29);
  \fill[LovartAccent] (1.85,0) rectangle (2.75,2.568);
  \draw[LovartMuted,line width=.4pt] (0,0)--(3.2,0);
  \node[anchor=south,text=gray!80!black,font=\sffamily\scriptsize\bfseries] at (0.9,1.31){0.21};
  \node[anchor=south,text=LovartAccent,font=\sffamily\scriptsize\bfseries] at (2.3,2.59){0.43};
  \node[anchor=north,align=center] at (0.9,-.05){Cycle\\$(i,\,i{+}18)$};
  \node[anchor=north,align=center] at (2.3,-.05){Adjacent\\$(2j,\,2j{+}1)$};
  \node[font=\sffamily\small\bfseries] at (1.6,3.75){(c) Tied-layer similarity};
\end{scope}
\end{tikzpicture} {}
  \caption{Controlled comparison of the RRT design choices with ours on Qwen3-4B, all trained with the same distillation recipe.
    (a)~Base-7 accuracy of cycle tying as used by RRT and of our adjacent tying at equal size (2.27B) and equal token budget.
    (b)~Base-7 accuracy of the static depth-wise LoRA used by RRT and of our input-conditioned gate on the main configuration.
    (c)~Mean cosine similarity between the residual updates of the two layers that each topology ties together, measured on the teacher.}
  \label{fig:rrt}
\end{figure}
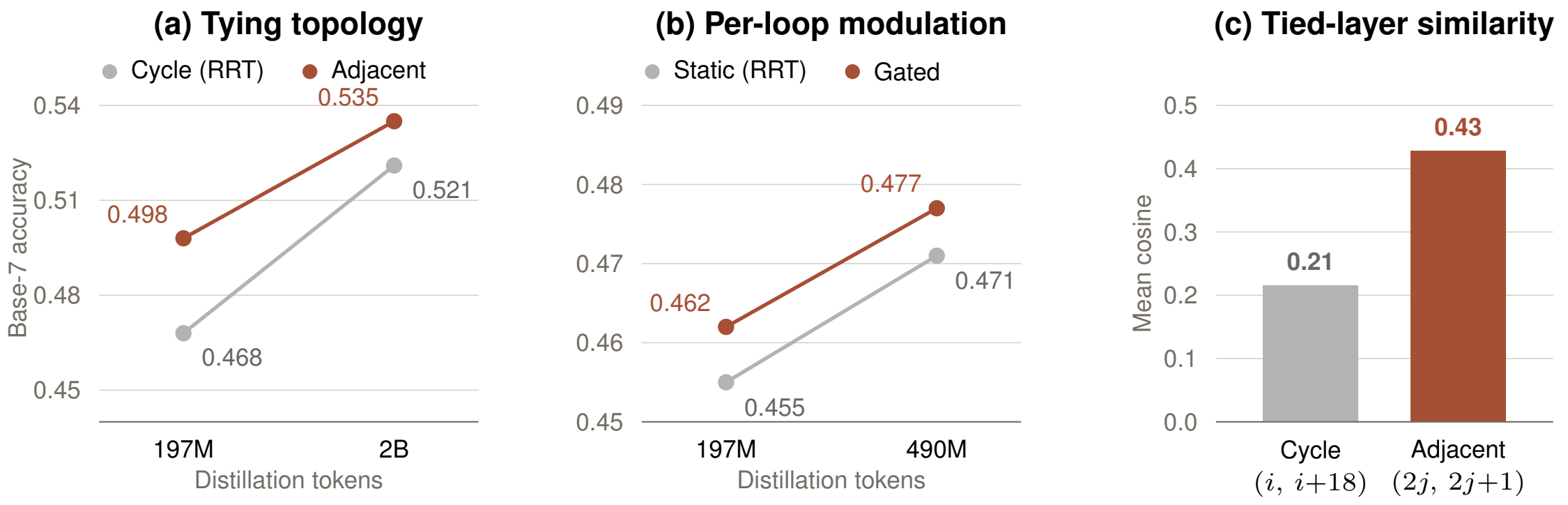

\paragraph{Tying topology.}
Figure~\ref{fig:rrt}a compares cycle tying with adjacent tying at the same parameter count of 2.27B, which is 57\% of the teacher.
Both models use the same static step-LoRA, so the only difference is which pair of layers shares a block.
With 197M distillation tokens, adjacent tying reaches 0.498 while cycle tying reaches 0.468, a gap of 4.7 points of teacher recovery.
When the budget grows to 2B tokens the ranking holds, with adjacent tying at 0.535 and cycle tying at 0.521, as listed in Table~\ref{tab:rrt} of the appendix.
Adjacent tying is the better topology at every budget we tested.

\paragraph{Similarity of tied layers.}
Figure~\ref{fig:rrt}c shows the similarity of the layers that each topology forces to share weights.
For every layer $i$ of the teacher we compute its residual update $u_i = h_i^{\text{out}} - h_i^{\text{in}}$ on held-out text and take the mean cosine similarity between the updates of the two tied layers.
Adjacent pairs have a mean cosine of 0.43, while the pairs $(i, i+18)$ chosen by cycle tying have a mean cosine of only 0.21.
A shared block under cycle tying must therefore play two roles that point in very different directions, and the rank-64 step correction has to absorb most of that difference.
Adjacent tying asks the same block to play two similar roles, which the shared backbone already captures and the step correction only needs to refine.
The topology result follows from this similarity gap, so we expect it to hold for other pretrained models in which neighboring layers are more alike than distant ones.

\paragraph{Per-loop modulation.}
Figure~\ref{fig:rrt}b compares the static depth-wise LoRA of RRT with our input-conditioned gate on the main configuration, in which all 36 layers are folded into adjacent pairs, using the data of Section~\ref{sec:exp-dense}.
The gate starts identical to the static variant and adds 0.08\% of the parameters.
At 197M tokens the gate improves base-7 from 0.455 to 0.462, and at 490M tokens it improves it from 0.471 to 0.477.
The gain is larger on generation, where at 490M tokens the gate raises GSM8K from 0.189 to 0.235 and MATH from 0.124 to 0.142, as listed in Table~\ref{tab:gate} of the appendix.
The distillation loss at the end of training is also lower with the gate at both budgets.
The trained gates leave the one-hot assignment for about 60\% of tokens at the second step, and for 30\% in the main model after 22B tokens.

\paragraph{Summary.}
Under the same training recipe, our design is better than RRT on both axes that distinguish the two methods.
Adjacent tying removes the mismatch between the two roles of a shared block, and the input-conditioned gate lets each token resolve whatever mismatch remains.
Because the static depth-wise LoRA of RRT is the one-hot special case of our gate, the second result also shows that the gate learns to leave that special case when given the chance.
For compression with a limited distillation budget, adjacent tying with an input-conditioned gate is the better way to convert a pretrained model into a looped one.

\subsection{Data Mixture Ablation}
\label{sec:exp-mix}

Table~\ref{tab:mix} compares three data mixtures for the same 2.27B fold of Qwen3-4B at a budget of about 20B tokens.
The first mixture combines general, mathematical and academic text in a ratio of 25:35:40.
The second, which we call the generic mixture, adds diverse web text drawn from many domains, with general, mathematical, academic and diverse text in a ratio of 15:30:20:35.
The third is the targeted mixture of Section~\ref{sec:setup}, which devotes a quarter of the data to multiple-choice questions from many subjects and includes the synthesized data.

\begin{table}[t]
  \centering
  \caption{Data mixture ablation for the 2.27B fold of Qwen3-4B.
    Every row is a separate run from the teacher weights with the same method and objective, and only the training data differs.
    The last column is the mean percentage of the teacher over the eight benchmarks.}
  \label{tab:mix}
  \small
  \setlength{\tabcolsep}{3.6pt}
  \renewcommand{\arraystretch}{1.15}
  \begin{tabularx}{\linewidth}{@{}Xrrrrrrrrrr@{}}
    \toprule
    \textbf{Mixture} & \textbf{Tokens} & \textbf{base-7} & \textbf{MMLU} &
      \textbf{Redux} & \textbf{Pro} & \textbf{GPQA} & \textbf{BBH} &
      \textbf{GSM8K} & \textbf{MATH} & \textbf{Avg.}\\
    \midrule
    General, math, academic & 20B & 0.590 & 0.533 & 0.566 & 0.238 & 0.283 & 0.332 & 0.718 & 0.408 & 75.6\%\\
    Generic, with diverse text & 22B & \textbf{0.592} & 0.549 & 0.584 & 0.252 & 0.288 & 0.364 & 0.740 & 0.438 & 78.6\%\\
    \rowcolor{LovartLight}
    \textbf{Targeted (ours)} & 22B & 0.580 & \textbf{0.556} & \textbf{0.587} & \textbf{0.301} & \textbf{0.308} & \textbf{0.400} & \textbf{0.788} & \textbf{0.498} & \textbf{83.6\%}\\
    \bottomrule
  \end{tabularx}
\end{table}

Adding diverse text raises every benchmark and lifts the average recovery from 75.6\% to 78.6\%, with the largest gains on BBH and MATH.
The targeted mixture adds another five points at the same budget of 22B tokens.
Its gains concentrate on the benchmarks that were furthest from the teacher, with MMLU-Pro rising by 4.9 points, BBH by 3.6, GSM8K by 4.8 and MATH by 6.0, while base-7 stays within about one point.
Together with the flattening of the scaling curve in Figure~\ref{fig:scaling}, this shows that once the budget is large enough, what the data covers decides which capabilities the folded model recovers, more than how much data it sees.

\subsection{Folding a Larger Teacher}
\label{sec:exp-scale}

To test whether Concertina transfers to a larger teacher, we fold Qwen3-8B with groups of $N=3$ adjacent layers into a model of 3.68B parameters, 45\% of the teacher.
It is distilled from Qwen3-8B on 22B tokens of general, mathematical, academic and diverse text in a ratio of 15:30:20:35.

\begin{table}[htbp]
  \centering
  \caption{Folding Qwen3-8B with groups of three layers.
    The last row gives the score of Concertina as a percentage of the teacher.}
  \label{tab:scale}
  \small
  \setlength{\tabcolsep}{4.5pt}
  \renewcommand{\arraystretch}{1.15}
  \begin{tabularx}{\linewidth}{@{}Xrrrrrrrrr@{}}
    \toprule
    \textbf{Model} & \textbf{Params} & \textbf{base-7} & \textbf{MMLU} &
      \textbf{Redux} & \textbf{Pro} & \textbf{GPQA} & \textbf{BBH} &
      \textbf{GSM8K} & \textbf{MATH}\\
    \midrule
    Qwen3-8B (teacher) & 8.19B & 0.660 & 0.726 & 0.768 & 0.462 & 0.369 & 0.553 & 0.895 & 0.587\\
    \rowcolor{LovartLight}
    \textbf{Concertina} ($N=3$) & 3.68B & 0.601 & 0.539 & 0.571 & 0.269 & 0.268 & 0.353 & 0.745 & 0.442\\
    \midrule
    \% of teacher & 45\% & 91\% & 74\% & 74\% & 58\% & 73\% & 64\% & 83\% & 75\%\\
    \bottomrule
  \end{tabularx}
\end{table}

Table~\ref{tab:scale} compares the folded model with Qwen3-8B on all eight benchmarks.
With less than half of the teacher's parameters, Concertina keeps 91\% of its commonsense score, 83\% of its GSM8K score and about three quarters of its MMLU, MMLU-Redux, GPQA and MATH scores.
Folding three layers into each shared block thus already recovers most of an 8B teacher, so the method is not specific to one model size or folding ratio.

\section{Discussion}
\label{sec:discussion}

\paragraph{Folding as an Experiment on Weight Organization.}
Because Concertina builds the student from the teacher's weights, it also tests whether two layers can be implemented by one set of weights, much like a lesion study in neuroscience.
The answer is that the redundancy between layers lies in their function rather than in their weights.
The difference between two adjacent layers is close to full rank and larger in norm than either layer, yet, as shown in Section~\ref{sec:exp-design}, the two layers write far more aligned updates into the residual stream than distant layers do.
\citet{lad2024remarkable} observed that middle layers can be deleted or swapped with little effect while the first and last layers are the most sensitive, and our reconstruction errors show the same profile, 20 to 200 times higher at the two ends of the network, as listed in Appendix~\ref{app:recon}.

\paragraph{Depth and Parameters.}
The truncated and folded students have almost the same size, so their difference separates the roles of depth and parameters.
As shown in Section~\ref{sec:exp-dense}, the folded model scores 8 to 11 points higher on the MMLU family and about 41 times higher on GSM8K, so at equal size keeping the depth helps every capability and generation the most.
When depth is kept, factual recall depends on parameters more than procedural skills do.
On SuperGPQA, both Concertina and the native Qwen3-1.7B lose more on recall questions than on calculation questions relative to the teacher, by 12.6 and 9.9 points respectively.
\citet{saunshi2025reasoning} observed a similar split for looped models trained from scratch, which reach worse perplexity than larger non-looped models but match them on reasoning.
These observations suggest that depth mainly supports procedures that can be applied repeatedly, while parameters mainly support facts that must be stored one by one.

\paragraph{Data Needs of an Inherited Student.}
For a student inherited from the teacher, additional data helps mainly when it exercises a capability that has not yet recovered.
As shown in Section~\ref{sec:exp-mix}, at a fixed budget of 22B tokens the targeted mixture raises the recovery of MMLU-Pro from 62\% to 75\% and also lifts BBH, GSM8K and MATH, while the scaling curve of the main model flattens after about 15B tokens.
A likely explanation is that the folded model keeps most of the teacher's knowledge but loses some of the paths that read it out, and each kind of data repairs the paths it exercises.
What the data covers therefore matters more than how much of it there is.

\paragraph{Evaluation of Compressed Models.}
Saturated commonsense benchmarks and perplexity overstate how much a compressed model retains, because likelihood-scored multiple choice does not require generation.
This is why the truncated model stays far above chance on them while failing on generation, and the finding that many layers can be removed with little loss~\citep{men2024shortgpt,gromov2024unreasonable} also rests mainly on such evaluation.
We therefore recommend that compressed language models also be evaluated on hard knowledge and generation tasks.

\paragraph{Limits of Compression.}
\citet{deletang2024language} showed that a language model is a strong general-purpose compressor.
This suggests that most of the information a model uses lies in its weights rather than in the words it reads.
Folding shows that even a fully trained model can be compressed further, but how far depends on what is compressed.
Skills resemble programs, which trade parameters for computation by running the same weights several times, while facts resemble random bits that each occupy their own capacity.
This agrees with \citet{kumar2025scaling}, who found that models pretrained on more data degrade more under quantization.
Measuring factual and procedural recovery separately across folding ratios would turn the broad claim that intelligence is compression into a question that experiments can answer.

\section{Conclusion}
\label{sec:conclusion}

We presented Concertina, which folds a pretrained language model into itself so that a compact student starts from the teacher's weights rather than from scratch.
Adjacent layers compute similar functions even though their weights differ, so one shared block with small corrections for each step and each token can closely approximate several layers.
Because the folded model keeps the depth of the teacher, it keeps multi-step skills such as mathematical reasoning, which a truncated model of the same size loses.
Because it also starts with most of the teacher's knowledge, distillation mainly needs data that exercises each skill rather than more data.
As a result, a 2.27B Concertina model recovers 84\% of Qwen3-4B on average from 22B tokens and outperforms Qwen3-1.7B, which is trained on over a thousand times more data.
More broadly, folding offers a cheap way to build small models from large ones, and a tool for studying what a model gains from depth and what it gains from parameters.

\clearpage
\phantomsection
\addcontentsline{toc}{section}{References}
\bibliographystyle{unsrtnat}
\bibliography{references}

\begin{thebibliography}{45}
\providecommand{\natexlab}[1]{#1}
\providecommand{\url}[1]{\texttt{#1}}
\expandafter\ifx\csname urlstyle\endcsname\relax
  \providecommand{\doi}[1]{doi: #1}\else
  \providecommand{\doi}{doi: \begingroup \urlstyle{rm}\Url}\fi

\bibitem[{Gemma Team}(2024)]{gemma2024gemma2}
{Gemma Team}.
\newblock Gemma 2: Improving open language models at a practical size.
\newblock \emph{arXiv preprint arXiv:2408.00118}, 2024.

\bibitem[{Qwen Team}(2025)]{yang2025qwen3}
{Qwen Team}.
\newblock Qwen3 technical report.
\newblock \emph{arXiv preprint arXiv:2505.09388}, 2025.

\bibitem[Bae et~al.(2025{\natexlab{a}})Bae, Fisch, Harutyunyan, Ji, Kim, and Schuster]{bae2025rrt}
Sangmin Bae, Adam Fisch, Hrayr Harutyunyan, Ziwei Ji, Seungyeon Kim, and Tal Schuster.
\newblock Relaxed recursive transformers: Effective parameter sharing with layer-wise {LoRA}.
\newblock In \emph{International Conference on Learning Representations}, 2025{\natexlab{a}}.
\newblock URL \url{https://arxiv.org/abs/2410.20672}.

\bibitem[Liu et~al.(2024{\natexlab{a}})Liu, Zeng, Li, Yan, Fu, Mei, and Chen]{liu2024foldgpt}
Songwei Liu, Chao Zeng, Lianqiang Li, Chenqian Yan, Lean Fu, Xing Mei, and Fangmin Chen.
\newblock {FoldGPT}: Simple and effective large language model compression scheme.
\newblock \emph{arXiv preprint arXiv:2407.00928}, 2024{\natexlab{a}}.

\bibitem[Hinton et~al.(2015)Hinton, Vinyals, and Dean]{hinton2015distilling}
Geoffrey Hinton, Oriol Vinyals, and Jeff Dean.
\newblock Distilling the knowledge in a neural network.
\newblock \emph{arXiv preprint arXiv:1503.02531}, 2015.

\bibitem[Kim and Rush(2016)]{kim2016sequence}
Yoon Kim and Alexander~M. Rush.
\newblock Sequence-level knowledge distillation.
\newblock In \emph{Proceedings of the Conference on Empirical Methods in Natural Language Processing}, pages 1317--1327, 2016.

\bibitem[Sanh et~al.(2019)Sanh, Debut, Chaumond, and Wolf]{sanh2019distilbert}
Victor Sanh, Lysandre Debut, Julien Chaumond, and Thomas Wolf.
\newblock {DistilBERT}, a distilled version of {BERT}: smaller, faster, cheaper and lighter.
\newblock \emph{arXiv preprint arXiv:1910.01108}, 2019.

\bibitem[Jiao et~al.(2020)Jiao, Yin, Shang, Jiang, Chen, Li, Wang, and Liu]{jiao2020tinybert}
Xiaoqi Jiao, Yichun Yin, Lifeng Shang, Xin Jiang, Xiao Chen, Linlin Li, Fang Wang, and Qun Liu.
\newblock {TinyBERT}: Distilling {BERT} for natural language understanding.
\newblock In \emph{Findings of the Association for Computational Linguistics: EMNLP}, pages 4163--4174, 2020.

\bibitem[Agarwal et~al.(2024)Agarwal, Vieillard, Zhou, Stanczyk, Ramos, Geist, and Bachem]{agarwal2024policy}
Rishabh Agarwal, Nino Vieillard, Yongchao Zhou, Piotr Stanczyk, Sabela Ramos, Matthieu Geist, and Olivier Bachem.
\newblock On-policy distillation of language models: Learning from self-generated mistakes.
\newblock In \emph{International Conference on Learning Representations}, 2024.

\bibitem[Gu et~al.(2024)Gu, Dong, Wei, and Huang]{gu2024minillm}
Yuxian Gu, Li~Dong, Furu Wei, and Minlie Huang.
\newblock {MiniLLM}: Knowledge distillation of large language models.
\newblock In \emph{International Conference on Learning Representations}, 2024.

\bibitem[Lu and {Thinking Machines Lab}(2025)]{lu2025onpolicy}
Kevin Lu and {Thinking Machines Lab}.
\newblock On-policy distillation.
\newblock Thinking Machines Lab: Connectionism (blog), 2025.
\newblock URL \url{https://thinkingmachines.ai/blog/on-policy-distillation/}.

\bibitem[Cloud et~al.(2026)Cloud, Le, Chua, Betley, Sztyber-Betley, Mindermann, Hilton, Marks, and Evans]{cloud2026language}
Alex Cloud, Minh Le, James Chua, Jan Betley, Anna Sztyber-Betley, S{\"o}ren Mindermann, Jacob Hilton, Samuel Marks, and Owain Evans.
\newblock Language models transmit behavioural traits through hidden signals in data.
\newblock \emph{Nature}, 652\penalty0 (8110):\penalty0 615--621, 2026.

\bibitem[Zhang et~al.(2026)Zhang, Jing, Zeng, Li, Wang, and Gong]{zhang2026post}
Ziyang Zhang, Yubin Jing, Yuanhao Zeng, Yuyao Li, Haofan Wang, and Yichen Gong.
\newblock Post-training leaves behavioral shadows on unrelated decisions.
\newblock \emph{arXiv preprint arXiv:2609.29233}, 2026.

\bibitem[Dehghani et~al.(2019)Dehghani, Gouws, Vinyals, Uszkoreit, and Kaiser]{dehghani2019universal}
Mostafa Dehghani, Stephan Gouws, Oriol Vinyals, Jakob Uszkoreit, and {\L}ukasz Kaiser.
\newblock Universal transformers.
\newblock In \emph{International Conference on Learning Representations}, 2019.

\bibitem[Lan et~al.(2020)Lan, Chen, Goodman, Gimpel, Sharma, and Soricut]{lan2020albert}
Zhenzhong Lan, Mingda Chen, Sebastian Goodman, Kevin Gimpel, Piyush Sharma, and Radu Soricut.
\newblock {ALBERT}: A lite {BERT} for self-supervised learning of language representations.
\newblock In \emph{International Conference on Learning Representations}, 2020.

\bibitem[Liu et~al.(2024{\natexlab{b}})Liu, Zhao, Iandola, Lai, Tian, Fedorov, Xiong, Chang, Shi, Krishnamoorthi, Lai, and Chandra]{liu2024mobilellm}
Zechun Liu, Changsheng Zhao, Forrest Iandola, Chen Lai, Yuandong Tian, Igor Fedorov, Yunyang Xiong, Ernie Chang, Yangyang Shi, Raghuraman Krishnamoorthi, Liangzhen Lai, and Vikas Chandra.
\newblock {MobileLLM}: Optimizing sub-billion parameter language models for on-device use cases.
\newblock In \emph{International Conference on Machine Learning}, pages 32431--32454, 2024{\natexlab{b}}.

\bibitem[Bae et~al.(2025{\natexlab{b}})Bae, Kim, Bayat, Kim, Ha, Schuster, Fisch, Harutyunyan, Ji, Courville, and Yun]{bae2025mor}
Sangmin Bae, Yujin Kim, Reza Bayat, Sungnyun Kim, Jiyoun Ha, Tal Schuster, Adam Fisch, Hrayr Harutyunyan, Ziwei Ji, Aaron Courville, and Se-Young Yun.
\newblock Mixture-of-recursions: Learning dynamic recursive depths for adaptive token-level computation.
\newblock In \emph{Advances in Neural Information Processing Systems}, 2025{\natexlab{b}}.

\bibitem[Nouriborji et~al.(2025)Nouriborji, Rohanian, and Rohanian]{nouriborji2025mol}
Mohammadmahdi Nouriborji, Morteza Rohanian, and Omid Rohanian.
\newblock Improving recursive transformers with mixture of {LoRAs}.
\newblock \emph{arXiv preprint arXiv:2512.12880}, 2025.

\bibitem[Mikaelyan et~al.(2025)Mikaelyan, Imani, Salvaris, Pathak, and Fayyaz]{mikaelyan2025deltallm}
Liana Mikaelyan, Ayyoob Imani, Mathew Salvaris, Parth Pathak, and Mohsen Fayyaz.
\newblock {DeltaLLM}: Compress {LLMs} with low-rank deltas between shared weights.
\newblock \emph{arXiv preprint arXiv:2501.18596}, 2025.

\bibitem[Smith et~al.(2025)Smith, Lin, Tuli, Jeelani, Gao, Shen, Jin, and Hsu]{smith2025flexigpt}
James~Seale Smith, Chi-Heng Lin, Shikhar Tuli, Haris Jeelani, Shangqian Gao, Yilin Shen, Hongxia Jin, and Yen-Chang Hsu.
\newblock {FlexiGPT}: Pruning and extending large language models with low-rank weight sharing.
\newblock In \emph{Proceedings of the Conference of the Nations of the Americas Chapter of the Association for Computational Linguistics}, pages 718--730, 2025.

\bibitem[Jaber and Jaber(2026)]{jaber2026ouroboros}
Jaber Jaber and Osama Jaber.
\newblock Ouroboros: Dynamic weight generation for recursive transformers via input-conditioned {LoRA} modulation.
\newblock \emph{arXiv preprint arXiv:2604.02051}, 2026.

\bibitem[Wang et~al.(2025)Wang, {\v{S}}iki{\'c}, Thiele, and Saukh]{wang2025modelfolding}
Dong Wang, Haris {\v{S}}iki{\'c}, Lothar Thiele, and Olga Saukh.
\newblock Forget the data and fine-tuning! just fold the network to compress.
\newblock In \emph{International Conference on Learning Representations}, 2025.

\bibitem[Men et~al.(2025)Men, Xu, Zhang, Yuan, Wang, Lin, Lu, Han, and Chen]{men2024shortgpt}
Xin Men, Mingyu Xu, Qingyu Zhang, Qianhao Yuan, Bingning Wang, Hongyu Lin, Yaojie Lu, Xianpei Han, and Weipeng Chen.
\newblock {ShortGPT}: Layers in large language models are more redundant than you expect.
\newblock In \emph{Findings of the Association for Computational Linguistics: ACL}, pages 20192--20204, 2025.

\bibitem[Gromov et~al.(2025)Gromov, Tirumala, Shapourian, Glorioso, and Roberts]{gromov2024unreasonable}
Andrey Gromov, Kushal Tirumala, Hassan Shapourian, Paolo Glorioso, and Daniel~A. Roberts.
\newblock The unreasonable ineffectiveness of the deeper layers.
\newblock In \emph{International Conference on Learning Representations}, 2025.

\bibitem[Muralidharan et~al.(2024)Muralidharan, Sreenivas, Joshi, Chochowski, Patwary, Shoeybi, Catanzaro, Kautz, and Molchanov]{muralidharan2024minitron}
Saurav Muralidharan, Sharath~Turuvekere Sreenivas, Raviraj Joshi, Marcin Chochowski, Mostofa Patwary, Mohammad Shoeybi, Bryan Catanzaro, Jan Kautz, and Pavlo Molchanov.
\newblock Compact language models via pruning and knowledge distillation.
\newblock In \emph{Advances in Neural Information Processing Systems}, 2024.

\bibitem[Hu et~al.(2022)Hu, Shen, Wallis, Allen-Zhu, Li, Wang, Wang, and Chen]{hu2022lora}
Edward~J. Hu, Yelong Shen, Phillip Wallis, Zeyuan Allen-Zhu, Yuanzhi Li, Shean Wang, Lu~Wang, and Weizhu Chen.
\newblock {LoRA}: Low-rank adaptation of large language models.
\newblock In \emph{International Conference on Learning Representations}, 2022.

\bibitem[Meng et~al.(2024)Meng, Wang, and Zhang]{meng2024pissa}
Fanxu Meng, Zhaohui Wang, and Muhan Zhang.
\newblock {PiSSA}: Principal singular values and singular vectors adaptation of large language models.
\newblock In \emph{Advances in Neural Information Processing Systems}, 2024.

\bibitem[Clark et~al.(2018)Clark, Cowhey, Etzioni, Khot, Sabharwal, Schoenick, and Tafjord]{clark2018think}
Peter Clark, Isaac Cowhey, Oren Etzioni, Tushar Khot, Ashish Sabharwal, Carissa Schoenick, and Oyvind Tafjord.
\newblock Think you have solved question answering? {T}ry {ARC}, the {AI2} reasoning challenge.
\newblock \emph{arXiv preprint arXiv:1803.05457}, 2018.

\bibitem[Zellers et~al.(2019)Zellers, Holtzman, Bisk, Farhadi, and Choi]{zellers2019hellaswag}
Rowan Zellers, Ari Holtzman, Yonatan Bisk, Ali Farhadi, and Yejin Choi.
\newblock {HellaSwag}: Can a machine really finish your sentence?
\newblock In \emph{Proceedings of the Annual Meeting of the Association for Computational Linguistics}, pages 4791--4800, 2019.

\bibitem[Bisk et~al.(2020)Bisk, Zellers, Le~Bras, Gao, and Choi]{bisk2020piqa}
Yonatan Bisk, Rowan Zellers, Ronan Le~Bras, Jianfeng Gao, and Yejin Choi.
\newblock {PIQA}: Reasoning about physical commonsense in natural language.
\newblock In \emph{Proceedings of the AAAI Conference on Artificial Intelligence}, pages 7432--7439, 2020.

\bibitem[Mihaylov et~al.(2018)Mihaylov, Clark, Khot, and Sabharwal]{mihaylov2018can}
Todor Mihaylov, Peter Clark, Tushar Khot, and Ashish Sabharwal.
\newblock Can a suit of armor conduct electricity? {A} new dataset for open book question answering.
\newblock In \emph{Proceedings of the Conference on Empirical Methods in Natural Language Processing}, pages 2381--2391, 2018.

\bibitem[Sakaguchi et~al.(2020)Sakaguchi, Le~Bras, Bhagavatula, and Choi]{sakaguchi2020winogrande}
Keisuke Sakaguchi, Ronan Le~Bras, Chandra Bhagavatula, and Yejin Choi.
\newblock {WinoGrande}: An adversarial {W}inograd schema challenge at scale.
\newblock In \emph{Proceedings of the AAAI Conference on Artificial Intelligence}, pages 8732--8740, 2020.

\bibitem[Paperno et~al.(2016)Paperno, Kruszewski, Lazaridou, Pham, Bernardi, Pezzelle, Baroni, Boleda, and Fern{\'a}ndez]{paperno2016lambada}
Denis Paperno, Germ{\'a}n Kruszewski, Angeliki Lazaridou, Ngoc~Quan Pham, Raffaella Bernardi, Sandro Pezzelle, Marco Baroni, Gemma Boleda, and Raquel Fern{\'a}ndez.
\newblock The {LAMBADA} dataset: Word prediction requiring a broad discourse context.
\newblock In \emph{Proceedings of the Annual Meeting of the Association for Computational Linguistics}, pages 1525--1534, 2016.

\bibitem[Hendrycks et~al.(2021{\natexlab{a}})Hendrycks, Burns, Basart, Zou, Mazeika, Song, and Steinhardt]{hendrycks2021mmlu}
Dan Hendrycks, Collin Burns, Steven Basart, Andy Zou, Mantas Mazeika, Dawn Song, and Jacob Steinhardt.
\newblock Measuring massive multitask language understanding.
\newblock In \emph{International Conference on Learning Representations}, 2021{\natexlab{a}}.

\bibitem[Gema et~al.(2025)Gema, Leang, Hong, Devoto, Mancino, Saxena, He, Zhao, Du, Madani, Barale, McHardy, Harris, Kaddour, van Krieken, and Minervini]{gema2025mmluredux}
Aryo~Pradipta Gema, Joshua Ong~Jun Leang, Giwon Hong, Alessio Devoto, Alberto Carlo~Maria Mancino, Rohit Saxena, Xuanli He, Yu~Zhao, Xiaotang Du, Mohammad Reza~Ghasemi Madani, Claire Barale, Robert McHardy, Joshua Harris, Jean Kaddour, Emile van Krieken, and Pasquale Minervini.
\newblock Are we done with {MMLU}?
\newblock In \emph{Proceedings of the Conference of the Nations of the Americas Chapter of the Association for Computational Linguistics}, pages 5069--5096, 2025.

\bibitem[Wang et~al.(2024)Wang, Ma, Zhang, Ni, Chandra, Guo, Ren, Arulraj, He, Jiang, Li, Ku, Wang, Zhuang, Fan, Yue, and Chen]{wang2024mmlupro}
Yubo Wang, Xueguang Ma, Ge~Zhang, Yuansheng Ni, Abhranil Chandra, Shiguang Guo, Weiming Ren, Aaran Arulraj, Xuan He, Ziyan Jiang, Tianle Li, Max Ku, Kai Wang, Alex Zhuang, Rongqi Fan, Xiang Yue, and Wenhu Chen.
\newblock {MMLU-Pro}: A more robust and challenging multi-task language understanding benchmark.
\newblock In \emph{Advances in Neural Information Processing Systems, Datasets and Benchmarks Track}, 2024.

\bibitem[Rein et~al.(2024)Rein, Hou, Stickland, Petty, Pang, Dirani, Michael, and Bowman]{rein2024gpqa}
David Rein, Betty~Li Hou, Asa~Cooper Stickland, Jackson Petty, Richard~Yuanzhe Pang, Julien Dirani, Julian Michael, and Samuel~R. Bowman.
\newblock {GPQA}: A graduate-level {G}oogle-proof {Q\&A} benchmark.
\newblock In \emph{Conference on Language Modeling}, 2024.

\bibitem[Suzgun et~al.(2023)Suzgun, Scales, Sch{\"a}rli, Gehrmann, Tay, Chung, Chowdhery, Le, Chi, Zhou, and Wei]{suzgun2023bbh}
Mirac Suzgun, Nathan Scales, Nathanael Sch{\"a}rli, Sebastian Gehrmann, Yi~Tay, Hyung~Won Chung, Aakanksha Chowdhery, Quoc~V. Le, Ed~H. Chi, Denny Zhou, and Jason Wei.
\newblock Challenging {BIG-Bench} tasks and whether chain-of-thought can solve them.
\newblock In \emph{Findings of the Association for Computational Linguistics: ACL}, pages 13003--13051, 2023.

\bibitem[Cobbe et~al.(2021)Cobbe, Kosaraju, Bavarian, Chen, Jun, Kaiser, Plappert, Tworek, Hilton, Nakano, Hesse, and Schulman]{cobbe2021gsm8k}
Karl Cobbe, Vineet Kosaraju, Mohammad Bavarian, Mark Chen, Heewoo Jun, Lukasz Kaiser, Matthias Plappert, Jerry Tworek, Jacob Hilton, Reiichiro Nakano, Christopher Hesse, and John Schulman.
\newblock Training verifiers to solve math word problems.
\newblock \emph{arXiv preprint arXiv:2110.14168}, 2021.

\bibitem[Hendrycks et~al.(2021{\natexlab{b}})Hendrycks, Burns, Kadavath, Arora, Basart, Tang, Song, and Steinhardt]{hendrycks2021math}
Dan Hendrycks, Collin Burns, Saurav Kadavath, Akul Arora, Steven Basart, Eric Tang, Dawn Song, and Jacob Steinhardt.
\newblock Measuring mathematical problem solving with the {MATH} dataset.
\newblock In \emph{Advances in Neural Information Processing Systems Datasets and Benchmarks Track}, 2021{\natexlab{b}}.

\bibitem[{M-A-P Team}(2025)]{du2025supergpqa}
{M-A-P Team}.
\newblock {SuperGPQA}: Scaling {LLM} evaluation across 285 graduate disciplines.
\newblock \emph{arXiv preprint arXiv:2502.14739}, 2025.

\bibitem[Lad et~al.(2025)Lad, Lee, Gurnee, and Tegmark]{lad2024remarkable}
Vedang Lad, Jin~Hwa Lee, Wes Gurnee, and Max Tegmark.
\newblock The remarkable robustness of {LLMs}: Stages of inference?
\newblock In \emph{Advances in Neural Information Processing Systems}, 2025.

\bibitem[Saunshi et~al.(2025)Saunshi, Dikkala, Li, Kumar, and Reddi]{saunshi2025reasoning}
Nikunj Saunshi, Nishanth Dikkala, Zhiyuan Li, Sanjiv Kumar, and Sashank~J. Reddi.
\newblock Reasoning with latent thoughts: On the power of looped transformers.
\newblock In \emph{International Conference on Learning Representations}, 2025.

\bibitem[Del{\'e}tang et~al.(2024)Del{\'e}tang, Ruoss, Duquenne, Catt, Genewein, Mattern, Grau-Moya, Wenliang, Aitchison, Orseau, Hutter, and Veness]{deletang2024language}
Gr{\'e}goire Del{\'e}tang, Anian Ruoss, Paul-Ambroise Duquenne, Elliot Catt, Tim Genewein, Christopher Mattern, Jordi Grau-Moya, Li~Kevin Wenliang, Matthew Aitchison, Laurent Orseau, Marcus Hutter, and Joel Veness.
\newblock Language modeling is compression.
\newblock In \emph{International Conference on Learning Representations}, 2024.

\bibitem[Kumar et~al.(2025)Kumar, Ankner, Spector, Bordelon, Muennighoff, Paul, Pehlevan, R{\'e}, and Raghunathan]{kumar2025scaling}
Tanishq Kumar, Zachary Ankner, Benjamin~F. Spector, Blake Bordelon, Niklas Muennighoff, Mansheej Paul, Cengiz Pehlevan, Christopher R{\'e}, and Aditi Raghunathan.
\newblock Scaling laws for precision.
\newblock In \emph{International Conference on Learning Representations}, 2025.

\end{thebibliography}

\clearpage
\appendix
\section{Reproduction of Prior Folding Methods}
\label{app:repro}

This appendix describes how we reproduce RRT and FoldGPT for the comparison in Section~\ref{sec:exp-rrt}.
Both baselines use the same FineWeb-Edu text, sequence length of 2048 tokens and evaluation harness as Concertina.

\paragraph{RRT.}
We follow the base uptraining recipe in Appendix E of \citet{bae2025rrt} which trains with cross-entropy only, and do not use the distillation variants that the work reports separately.
The model ties layer $i$ with layer $i+18$, initializes each shared block as the average of the two tied layers, and adds an SVD-initialized rank-64 LoRA to the seven linear layers of every loop.
Norms are tied across loops.
We train with AdamW at a learning rate of $2\times10^{-4}$ with cosine decay, 200 warmup steps and a weight decay of 0.1.
The 197M point is a complete run of 6000 steps with 32,768 tokens per step.
The 0.49B, 1B and 2B points are snapshots of a single run with a 20B-token cosine schedule and 131,072 tokens per step, taken at steps 3740, 7630 and 15259.

\paragraph{FoldGPT.}
We follow \citet{liu2024foldgpt} in three steps.
First, a learnable gate $g(a)=a^2/(a^2+\epsilon)$ is attached to every attention and feed-forward module, and only the gates are trained for 1000 steps on cross-entropy plus a penalty weighted by the FLOPs share of each module, with $\lambda=0.1$ and a learning rate of $2\times10^{-3}$.
This removes blocks 12, 13, 15, 16, 20 and 21, which agrees with the redundancy of middle layers reported by \citet{men2024shortgpt}.
Second, a dynamic program selects 12 pairs of adjacent remaining blocks with the lowest total gate importance, and within each pair the more important block becomes the parent.
The child block is $\mathrm{diag}(S)$ times the parent weight, with $S$ initialized by row-wise least squares, and keeps its own norm.
Blocks 4, 5, 6, 7, 30 and 31 stay unshared, which leaves 18 unique blocks and 2.21B parameters.
Third, the backbone is frozen and only a rank-8 LoRA on every parent, the scales and the norms are trained, about 8.8M parameters in total, with the same data, loss and schedule as RRT and no weight decay.
We drop the child bias so that the result exports as a standard Qwen3 model, and we omit the last-layer distillation term, whose weight in the original work is $10^{-5}$.
We also replace the original 50K Alpaca instructions with FineWeb-Edu and raise the learning rate from $10^{-5}$ to $2\times10^{-4}$ to match the RRT setting.
The exported model matches the training-time model with a maximum logit difference of $3\times10^{-4}$ in FP32 and identical top-1 predictions.

\begin{table}[htbp]
  \centering
  \caption{Per-benchmark scores of the prior folding methods and Concertina on Qwen3-4B.
    Every model has 18 unique blocks, which gives 2.21B parameters for FoldGPT, 2.34B for RRT and 2.27B for Concertina.
    Each task is evaluated on at most 1000 examples and OpenBookQA on 500.}
  \label{tab:baselines-full}
  \small
  \setlength{\tabcolsep}{5pt}
  \renewcommand{\arraystretch}{1.15}
  \begin{tabularx}{\linewidth}{@{}Xrrrrrrrr@{}}
    \toprule
    \textbf{Method} & \textbf{Tokens} & \textbf{Base-7} & \textbf{MMLU} &
      \textbf{Redux} & \textbf{Pro} & \textbf{SuperGPQA} & \textbf{11-task} & \textbf{PPL}\\
    \midrule
    FoldGPT & 197M & 0.410 & 0.247 & 0.258 & 0.133 & 0.093 & 0.327 & 23.1\\
    FoldGPT & 0.49B & 0.419 & 0.234 & 0.234 & 0.110 & 0.080 & 0.327 & 21.2\\
    FoldGPT & 1B & 0.426 & 0.224 & 0.228 & 0.111 & 0.085 & 0.330 & 20.3\\
    FoldGPT & 2B & 0.430 & 0.238 & 0.242 & 0.113 & 0.083 & 0.335 & 19.8\\
    \midrule
    RRT & 197M & 0.415 & 0.230 & 0.229 & 0.122 & 0.103 & 0.326 & 17.5\\
    RRT & 0.49B & 0.457 & 0.256 & 0.269 & 0.106 & 0.111 & 0.358 & 15.6\\
    RRT & 1B & 0.475 & 0.247 & 0.261 & 0.119 & 0.099 & 0.368 & 14.0\\
    RRT & 2B & 0.480 & 0.265 & 0.271 & 0.114 & 0.092 & 0.373 & 12.6\\
    \midrule
    \rowcolor{LovartLight}
    Concertina & 197M & 0.498 & 0.259 & 0.287 & 0.106 & 0.100 & 0.385 & 15.2\\
    \bottomrule
  \end{tabularx}
\end{table}

Table~\ref{tab:rrt} lists the scores of the tying topology ablation of Section~\ref{sec:exp-design}.
Both variants use static rank-64 step-LoRA and the distillation objective of Section~\ref{sec:method-kd} on FineWeb-Edu, and are evaluated with the base-7 protocol of Table~\ref{tab:baselines-full}, on which the Qwen3-4B teacher scores 0.631.

\begin{table}[htbp]
  \centering
  \caption{Controlled comparison of cycle tying as used by RRT with adjacent tying on Qwen3-4B.
    Both variants use static step-LoRA and keep all other settings fixed.
    Scores are base-7 averages on at most 1000 examples per task, and the percentage is relative to the Qwen3-4B teacher at 0.631.}
  \label{tab:rrt}
  \small
  \setlength{\tabcolsep}{6pt}
  \renewcommand{\arraystretch}{1.15}
  \begin{tabularx}{\linewidth}{@{}Xrrrrr@{}}
    \toprule
    \textbf{Variant} & \textbf{Params} & \textbf{Tokens} &
      \textbf{Base-7} & \textbf{\% teacher} & \textbf{$\Delta$}\\
    \midrule
    Cycle tying, RRT & 2.27B & 197M & 0.468 & 74.2\% & \\
    \rowcolor{LovartLight}
    Adjacent tying, ours & 2.27B & 197M & \best{0.498} & \best{78.9\%} & $+0.030$\\
    \midrule
    Cycle tying, RRT & 2.27B & 2B & 0.521 & 82.6\% & \\
    \rowcolor{LovartLight}
    Adjacent tying, ours & 2.27B & 2B & \best{0.535} & \best{84.8\%} & $+0.014$\\
    \bottomrule
  \end{tabularx}
\end{table}

Table~\ref{tab:gate} gives the full evaluation of the per-loop modulation ablation of Section~\ref{sec:exp-design}, on the main configuration with all 36 layers folded into adjacent pairs.
All runs use the same data, seed and schedule and are fully annealed at each budget, and all eight benchmarks are scored on their full evaluation sets as in Section~\ref{sec:setup}.
At 197M tokens the MMLU family, GPQA and BBH are still at or below chance, so base-7, GSM8K and MATH are the informative columns there.
The third row gives the first step a trainable correction initialized with a random $A_1$ and a zero $B_1$, as RRT does for a zero difference, so that the gate mixes two nonzero corrections.

\begin{table}[htbp]
  \centering
  \caption{Full evaluation of the static depth-wise LoRA and the input-conditioned gate on the main configuration of Qwen3-4B with 2.27B parameters.}
  \label{tab:gate}
  \small
  \setlength{\tabcolsep}{4pt}
  \renewcommand{\arraystretch}{1.15}
  \begin{tabularx}{\linewidth}{@{}Xrrrrrrrrr@{}}
    \toprule
    \textbf{Variant} & \textbf{Tokens} & \textbf{base-7} & \textbf{MMLU} & \textbf{Redux} & \textbf{Pro} & \textbf{GPQA} & \textbf{BBH} & \textbf{GSM8K} & \textbf{MATH}\\
    \midrule
    Static depth-wise LoRA & 197M & 0.455 & 0.261 & 0.260 & 0.113 & 0.242 & 0.234 & 0.106 & 0.077\\
    \rowcolor{LovartLight}
    Input-conditioned gate & 197M & 0.462 & 0.238 & 0.234 & 0.115 & 0.247 & 0.245 & 0.132 & 0.074\\
    Gate, trainable first step & 197M & 0.462 & 0.259 & 0.260 & 0.118 & 0.237 & 0.246 & 0.140 & 0.085\\
    \midrule
    Static depth-wise LoRA & 490M & 0.471 & 0.254 & 0.254 & 0.116 & 0.258 & 0.236 & 0.189 & 0.124\\
    \rowcolor{LovartLight}
    Input-conditioned gate & 490M & \textbf{0.477} & \textbf{0.262} & \textbf{0.267} & 0.116 & \textbf{0.278} & 0.235 & \textbf{0.235} & \textbf{0.142}\\
    \bottomrule
  \end{tabularx}
\end{table}

\FloatBarrier
\section{Per-Task Commonsense Scores}
\label{app:base7}

Table~\ref{tab:base7} lists the seven tasks that make up the base-7 average for the models in Sections~\ref{sec:exp-main}, \ref{sec:exp-dense} and~\ref{sec:exp-scale}.
All seven tasks are scored zero-shot by log-likelihood on their full evaluation sets.
ARC-Challenge, ARC-Easy, HellaSwag and OpenBookQA use length-normalized accuracy, and PIQA, WinoGrande and LAMBADA use plain accuracy, with LAMBADA in its OpenAI version.
The base-7 score is the unweighted mean of the seven tasks.
On the 4B scale, the main model keeps at least 95\% of the teacher on ARC-Easy, PIQA, OpenBookQA and WinoGrande, and it scores above Qwen3-1.7B on six of the seven tasks.
Under the same 20B-token distillation recipe, Concertina also scores above the truncated dense model on six of the seven tasks.

\begin{table}[htbp]
  \centering
  \caption{Per-task zero-shot scores behind the base-7 average.
    The 4B rows correspond to Tables~\ref{tab:main} and~\ref{tab:dense}, where both 20B models share the recipe of Section~\ref{sec:exp-dense}, and the 8B rows correspond to Table~\ref{tab:scale}.
    The second header row gives the metric of each task, where norm denotes length-normalized accuracy.}
  \label{tab:base7}
  \small
  \setlength{\tabcolsep}{3.2pt}
  \renewcommand{\arraystretch}{1.15}
  \begin{tabularx}{\linewidth}{@{}>{\raggedright\arraybackslash}X*{8}{c}@{}}
    \toprule
    \textbf{Model} & \textbf{ARC-C} & \textbf{ARC-E} & \textbf{HellaSwag} &
      \textbf{PIQA} & \textbf{OBQA} & \textbf{WinoGrande} & \textbf{LAMBADA} & \textbf{base-7}\\
    & \footnotesize norm & \footnotesize norm & \footnotesize norm & \footnotesize acc & \footnotesize norm & \footnotesize acc & \footnotesize acc & \footnotesize mean\\
    \midrule
    Qwen3-4B (teacher) & 0.543 & 0.785 & 0.676 & 0.750 & 0.402 & 0.659 & 0.594 & 0.630\\
    Qwen3-1.7B & 0.431 & 0.709 & 0.590 & 0.726 & 0.362 & 0.611 & 0.506 & 0.562\\
    Truncated dense (20B) & 0.445 & 0.729 & 0.577 & 0.734 & 0.404 & 0.620 & 0.482 & 0.570\\
    \rowcolor{LovartLight}
    Concertina (20B) & 0.463 & 0.755 & 0.614 & 0.745 & 0.396 & 0.645 & 0.515 & 0.590\\
    \rowcolor{LovartLight}
    Concertina (22B, main) & 0.458 & 0.746 & 0.585 & 0.731 & 0.382 & 0.646 & 0.511 & 0.580\\
    \quad \textit{\% of teacher} & 84\% & 95\% & 86\% & 97\% & 95\% & 98\% & 86\% & 92\%\\
    \midrule
    Qwen3-8B (teacher) & 0.567 & 0.819 & 0.734 & 0.768 & 0.410 & 0.677 & 0.643 & 0.660\\
    \rowcolor{LovartLight}
    Concertina ($N=3$) & 0.446 & 0.735 & 0.609 & 0.744 & 0.412 & 0.669 & 0.596 & 0.601\\
    \bottomrule
  \end{tabularx}
\end{table}

\FloatBarrier
\section{Reconstruction Error of Adjacent Layer Pairs}
\label{app:recon}

Table~\ref{tab:recon} measures how well one shared block can stand in for each pair of adjacent layers of Qwen3-4B before any training.
For each pair we build a folded block with rank-32 SVD-initialized step corrections, feed it the hidden states that the teacher produces at the input of the pair, and compare its output with the output of the two original layers.
We report the mean squared error normalized by the variance of the original output.
The middle pairs from layer 6 to layer 21 have errors between 0.008 and 0.018, the error rises steadily over the later pairs, and the first three pairs and the last pair are 20 to 200 times harder to fold than the middle.

\begin{table}[htbp]
  \centering
  \caption{Normalized reconstruction error of each adjacent layer pair of Qwen3-4B under an untrained fold with rank-32 step corrections.
    Each column is one pair of layers, indexed from zero.}
  \label{tab:recon}
  \small
  \setlength{\tabcolsep}{4pt}
  \renewcommand{\arraystretch}{1.15}
  \begin{tabular*}{\linewidth}{@{}l@{\extracolsep{\fill}}*{9}{c}@{}}
    \toprule
    \textbf{Layer pair} & $(0,1)$ & $(2,3)$ & $(4,5)$ & $(6,7)$ & $(8,9)$ & $(10,11)$ & $(12,13)$ & $(14,15)$ & $(16,17)$\\
    NMSE & 1.963 & 0.335 & 0.387 & 0.0083 & 0.0127 & 0.0103 & 0.0121 & 0.0177 & 0.0088\\
    \midrule
    \textbf{Layer pair} & $(18,19)$ & $(20,21)$ & $(22,23)$ & $(24,25)$ & $(26,27)$ & $(28,29)$ & $(30,31)$ & $(32,33)$ & $(34,35)$\\
    NMSE & 0.0092 & 0.0109 & 0.0257 & 0.0328 & 0.0440 & 0.0684 & 0.0930 & 0.130 & 0.492\\
    \bottomrule
  \end{tabular*}
\end{table}

\FloatBarrier
\section{Limitations}
\label{sec:limitations}

Because each shared block executes $N$ loop steps, folding reduces parameter storage, which benefits memory-constrained devices, but it keeps the forward FLOPs and the KV cache of the full-depth teacher rather than reducing them as width or depth pruning does.
In addition, our experiments cover the Qwen3 family at the 4B and 8B scales and do not test other model families or wall-clock latency across serving engines.
MATH is evaluated on a subset of 1{,}250 problems, and our hardest knowledge benchmarks are MMLU-Pro and GPQA-diamond, the latter with only 198 questions.
These limits mark where further systems work, data and evaluation would most improve the conclusions.

\end{document}